\documentclass{article}

\usepackage{PRIMEarxiv}

\usepackage[utf8]{inputenc} 
\usepackage[T1]{fontenc}    
\usepackage{hyperref}       
\usepackage{url}            
\usepackage{booktabs}       
\usepackage{amsfonts}       
\usepackage{nicefrac}       
\usepackage{microtype}      
\usepackage{lipsum}
\usepackage{graphicx}
\graphicspath{{media/}}     

\usepackage{dsfont}

\usepackage{algorithm}
\usepackage{algorithmic}

\usepackage{setspace}

\usepackage{physics}
\usepackage{amsmath}
\usepackage{tikz}
\usepackage{mathdots}
\usepackage{yhmath}
\usepackage{cancel}
\usepackage{color}
\usepackage{siunitx}
\usepackage{array}
\usepackage{multirow}
\usepackage{amssymb}
\usepackage{gensymb}
\usepackage{tabularx}
\usepackage{extarrows}
\usepackage{booktabs}
\usetikzlibrary{fadings}
\usetikzlibrary{patterns}
\usetikzlibrary{shadows.blur}
\usetikzlibrary{shapes}
\usetikzlibrary{positioning}

\title{Company2Vec - German Company Embeddings \\ based on Corporate Websites
\thanks{
Preprint of an article published in International Journal of Information Technology \& Decision Making © copyright World Scientific Publishing Company: https://www.worldscientific.com/worldscinet/ijitdm.\\
\textit{\underline{Citation}}: 
\textbf{C. Gerling, Company2Vec - German Company Embeddings based on
Corporate Websites, International Journal of Information Technology \& Decision Making (2023). DOI:10.1142/S0219622023500694.}} 
}

\author{
  Christopher Gerling \\
  Chair of Information Systems \\ 
  Humboldt University of Berlin\\  
  Berlin, Germany \\
  \texttt{gerlingc@hu-berlin.de}
}

\begin{document}
\maketitle

\begin{abstract}
With Company2Vec, the paper proposes a novel application in representation learning. The model analyzes business activities from unstructured company website data using Word2Vec and dimensionality reduction. Company2Vec maintains semantic language structures and thus creates efficient company embeddings in fine-granular industries. These semantic embeddings can be used for various applications in banking.

Direct relations between companies and words allow semantic business analytics (e.g. top-n words for a company). Furthermore, industry prediction is presented as a supervised learning application and evaluation method. The vectorized structure of the embeddings allows measuring companies similarities with the cosine distance. Company2Vec hence offers a more fine-grained comparison of companies than the standard industry labels (\textit{NACE}). This property is relevant for unsupervised learning tasks, such as clustering. An alternative industry segmentation is shown with k-means clustering on the company embeddings. Finally, this paper proposes three algorithms for (1) \textit{firm-centric}, (2) \textit{industry-centric} and (3) \textit{portfolio-centric} peer-firm identification.
\end{abstract}


\section{Introduction}\label{Sec:intro}
The vectorized representation of objects has been subject of research in computer science for many years. Especially in the field of computer linguistics, methods have been developed that can efficiently represent words and texts in a vector space. A widely-used method for unstructured (textual) data transformation is Word2Vec. \cite{mikolov2013efficient} Due to its generic architecture, this procedure is applied to a variety of other problems. Although more powerful language models have evolved in recent years, Word2Vec remains a state-of-the-art solution for creating generalized object embeddings.

The efficient representation of any object in a vector space has been shown for documents \cite{le2014distributed}, products \cite{gabelP2V} and music \cite{music2vec}, for instance. These generalized embeddings combine context information and thus allow a direct numerical comparison between two objects of the same type. Object embeddings can also be used as a structured input for any machine learning model.

This paper proposes a novel application of representation learning in the area of companies and industry structures. Until now, companies have been classified into 21 different industries with multiple sub-industries according to the \textit{NACE} code. \cite{Destatis2008} However, this distinct classification is too narrow considering companies with more than one business activity. Subtle differences between enterprises from the same industry sector are thus neglected.

A vectorized representation of companies opens up the possibility of representing these fine-granular differences quantitatively. Hence, similarities between industries or companies can be measured. Generalized company embeddings are useful in numerous processes of a bank: 

Automatically generated embeddings can first of all be used to obtain descriptions about business objectives of firms (topic modeling). Then, new customer acquisition or sustainable supplyer selection \cite{Gulsen} is supported by identifying similarities between companies. In addition, products demanded by certain companies can be recommended to similar firms. Furthermore, company embeddings are also helpful from a risk perspective: The fine-granular vector representations contain additional information for credit scoring. Finally, company portfolios are easily evaluated in terms of diversification and cluster risk.  \cite{Gulsen}

This paper contributes to literature as follows: 
First, a method is presented for creating German company embeddings using Word2Vec on company webpages (a subpart of complete websites). This method (\textit{Company2Vec}) uses unstructured textual and visual data from almost 42k company webpages. The evaluation process of this initial step identifies the most accurate model for numerical company representation and company similarity estimation.

Secondly, the resulting company embeddings are evaluated according to semantic criteria in order to show empirical evidence of the presented transformation.

Finally, the example of efficient peer-firm identification for various purposes gives insights into possible applications. Efficiency is assessed in terms of accuracy and complexity. The objective is to develop three low-complexity recommender algorithms for \textit{firm-centric}, \textit{industry-centric} and \textit{portfolio-centric} peer firm identification.

\section{Literature Review}\label{Sec:literature-review}
Research in company representation learning is concerned with two main objectives: (1) supervised classification methods for \textit{industry prediction} and (2) developing \textit{unsupervised learning methods} for e.g. alternative company segmentation or peer-firm identification. Both literature streams provide an important theoretical foundation for Company2Vec. 

Investigations with unstructured data have already been carried out for the purpose of industry classification. The findings of supervised industry prediction are used to assess the company representations in a detailed evaluation. The current state of research is presented in subsection \ref{Sec:IndustryPrediction}.

Concerning unsupervised learning, company similarities are measured and used for multiple applications. The process to form alternative segments of similar firms (\ref{Sec:Segmentation}) or to identify peer firms (\ref{Sec:PFI}) has been carried out in various studies.


\subsection{Industry Prediction (Supervised Learning)} \label{Sec:IndustryPrediction}
The use of textual (unstructured) data for company industry prediction has been an especially challenging research task for many years. Research in the field aims at optimising supervised classification models for industry prediction based on company-specific features. Pierre \cite{Pierre2001OnTA} highlights that ``\textit{in practice even human assigned classifications may only achieve scores between 0.7 and 0.9, depending on the} [multiclass] \textit{classification task. This is because to some extent classification is a subjective task and there are usually 'grey areas' in a classification scheme}". For that reason, the multiclass classification evaluation is often softened by assessing the model based on the \textit{Top-N} industry predictions (N-vs-Rest) instead of One-vs-Rest. Pierre \cite{Pierre2001OnTA} conducts the first industry classification study on textual data from 29,998 company web domains. However, the bag-of-words approach fails to catch information on large texts. In 2015, Berardi et al. \cite{BerardiEtAl2015} classify 20,000 company websites with enhanced analytical methods. The authors achieve accuracy scores close to human labeling accuracies using term frequency–inverse document frequency (TF-IDF) for website representation and a support vector machine (SVM) classifier.

Apart from company websites, Bernstein et al. \cite{BernsteinEtAl2003} use business newspaper corpora and count the number of companies' co-occurrences in articles under the assumption that those must be related to each other. A relational vector-space model (RVS) is able to classify 65\% of firms correctly into 12 industry sectors. The authors address problems in precisely identifying rare classes resulting from severe industry class imbalance.

\subsection{Unsupervised Company Learning}
 
\subsubsection{Company Segmentation} \label{Sec:Segmentation}
Literature on unsupervised learning in company similarity research focuses primarily on building sensible clusters of similar firms. In a majority of cases, the intention is to propose alternative industry labels that differ from the official NACE classification scheme. The main reason for this are certain limitations of existing classification systems. \cite{Kee2019PeerFI} Fodor \cite{Fodor2014FormingSG} argues that clustering firms based on their financial structure, instead of clustering based on business models, can provide additional and more accurate information in stock market prediction. 

Furthermore, Fang et al. \cite{Fang2013LDABasedIC} state that predefined classification systems are stable and ``\textit{cannot capture the dynamic aspect of the industry}". Hence many researchers have developed alternative industry classes by performing cluster analyses on company data. The results strongly depend on the collected input data and on the target use case. For instance, in the study of 2013, Fang et al. \cite{Fang2013LDABasedIC} propose a number of 60 clusters, whereas Fodor \cite{Fodor2014FormingSG} later identifies 25 clusters of which only 10 were interpretable. Moreover, Gkotsis et al. \cite{GkotsiEtAl2018} use hierarchical clustering and find 5 macro-clusters that can be subdivided into many smaller segments depending on the application. 

The most recent development in unsupervised company learning suggests dimensionality reduction algorithms for refining the embedding space and reducing computational complexity: Husmann et al. \cite{HUSMANN2022116598} use high-dimensional financial company data and apply t-distributed stochastic neighbor embedding (t-SNE). In combination with spectral clustering, Husmann et al. \cite{HUSMANN2022116598} create company segments that are used for company classification and financial performance analysis.

\subsubsection{Peer Firm Identification} \label{Sec:PFI}
Since clustering algorithms are based on distance measures between firms, an adequate numerical company representation is a crucial step in industry segmentation. In the respective literature, the numerical distance between companies forms a basis for \textit{peer firm identification}. Some authors focus on \textbf{structured data} for determining distances between firms: Fodor \cite{Fodor2014FormingSG} analyses common size financial statements consisting of income statements and ratios of balance sheets. Similar to this approach, Gkotsis et al. \cite{GkotsiEtAl2018} use quantitative information from the US patent register as a proxy to differentiate firms by their technological capabilities.

In addition to these static quantitative approaches, some human-centred studies have been carried out: Lee at al. \cite{LEE2015410} indirectly model companies' similarities by analysing the online traffic on the US company register EDGAR. The presented algorithm identifies co-searches of multiple firms by a single person under the assumption that these firms must show a positive relation to each other. More directly, Kaustia\&Rantala \cite{kaustia_rantala_2020} combine estimations of multiple financial analysts in order to model the relatedness between firms. 

Focusing on \textbf{unstructured data}, Rönnqvist\&Sarlin \cite{RoennqvistEtAl2015} and Hoberg\&Phillips \cite{HobergEtAl2016} use textual data on a very basic analytical level. Following the original idea of Bernstein et al. \cite{BernsteinEtAl2003}, the study by Rönnqvist\&Sarlin \cite{RoennqvistEtAl2015} measures the relatedness of banks by counting the co-occurrences of banks' names in newspaper articles. Hoberg et al. \cite{HobergEtAl2016} instead use product descriptions from annual reports for a high-dimensional TF-IDF vector representation of each company. At that time, more sophisticated models were already available.

Recent advances in analysing unstructured data reveal new possibilities in identifying peer firms. \textbf{Natural language processing} (NLP) provides different tools to show similarities between firms. In that manner, topic modeling is the key aspect for understanding firms' business activities. Fang et al. \cite{Fang2013LDABasedIC} present the first advanced NLP model for calculating company similarities. Fang et al. \cite{Fang2013LDABasedIC} base their topic modeling on firms' business descriptions from annual reports (K-10 forms). In particular, a latent Dirichlet allocation (LDA) model is applied. In their study design, Fang et al. \cite{Fang2013LDABasedIC} differentiate between two analytical use cases: In a \textit{firm-centric} perspective, the model begins with one single firm and generates a list of respective peer firms. Moreover, an \textit{industry-centric} perspective uses clustering for identifying firms operating in the same industry segment. 

A modern state-of-the-art solution for peer firm identification is presented by Kee \cite{Kee2019PeerFI}. This study is the first to suggest using word embeddings. Kee \cite{Kee2019PeerFI} uses a Word2Vec model to train word embeddings. The training corpus consists of a 10-year collection of 21 million English newspaper articles from Reuters. During the training process, the model implicitly learned the context in which certain company names appear. The word vector of the company name is then used as the final company embedding. \cite{Kee2019PeerFI} Determining the cosine distance between two vector representations of firms provides a parameter describing their similarity.

\subsection{Unsolved Problems}

In summary, the majority of recent studies refer to financial data or basic textual BOW/TF-IDF representations for companies. Current developments in NLP highlight promising potentials of vectorized company representation. New applications are a crucial challenge for older and much simpler methods. To address these complex applications comprehensively, more complex models are necessary. Thus, these models will be further examined in the work ahead.

\paragraph{German Companies.}
First of all, the presented studies on NLP use data from mostly North American companies. Thus, textual analyses were performed on English text corpora. Hence one lack of research is concerned with studies on German company embeddings. 

\paragraph{Data Collection.}
Pierre \cite{Pierre2001OnTA} and Berardi et al. \cite{BerardiEtAl2015} were able to build strong classification models based on official company webpage data. 
Assuming that the affiliation to a certain industry segment can be used as a proxy for firms' similarities, this data source has proved the most promising results so far. However, Pierre \cite{Pierre2001OnTA} and Berardi et al. \cite{BerardiEtAl2015} solely focus on textual data from company webpages. State-of-the-art webpages also offer additional information for data collection. Especially visual data from images on company websites have not been in the focus of past studies and will be included in this work.

\paragraph{Data Transformation.}
Taking a methodological perspective, the ideas presented by Fang et al. \cite{Fang2013LDABasedIC} and Kee \cite{Kee2019PeerFI} are crucial for the design of this study. In particular, the methods used to transform companies into numerical representations will be a key aspect of this work.
On a side note, the Word2Vec model by Kee \cite{Kee2019PeerFI} offers an even more recent algorithm for NLP than the LDA model by Fang et al. \cite{Fang2013LDABasedIC}. 

However, the Word2Vec model by Kee \cite{Kee2019PeerFI} has solely been trained on a textual newspaper corpus. One major drawback in this setting is the number of appearances of companies in the newspaper text corpus. Smaller companies will rarely be present in a set of newspaper articles. Consequently, the word vector might not be able to represent the company correctly: A small number of appearances will result in biased vectors. 

Furthermore, depending on the window size of the Word2Vec model, company names tend to appear in similar sentence structures or contexts. The cosine similarity, which can geometrically be interpreted both as the angle between two vectors or as the correlation between them, can hardly represent company similarities in a full range of value with $[-1;+1]$. This complicates the interpretation of similarity scores. This is an additional lack of research that has not been addressed so far.

\paragraph{Identifying Peer Firms.}
A numerical company representation as proposed by Fang et al. \cite{Fang2013LDABasedIC} and Kee \cite{Kee2019PeerFI} offers multiple possibilities for peer firm identification.
While Fang et al. \cite{Fang2013LDABasedIC} only distinguish between \textit{firm-centric} and \textit{industry-centric} perspectives, a \textit{portfolio-centric} analysis has yet to be developed. With this additional perspective, peer firms for a firm portfolio with multiple companies could be identified. The ability to create semantic combinations of firms in a portfolio might be a fruitful addition.

\section{Empirical Data}\label{Sec:experiment}

\subsection{Data Sets}

\subsubsection{Company Information}
The original data set contains the following variables: \textit{company name}, \textit{URL} and the corresponding \textit{industry classes}. Considering the industry classes, the data set distinguishes two granularities for evaluation. The NACE codes offer several hierarchical levels of industry classification. The \textit{level-1} granularity distinguishes between 21 distinct industry sections (A-U) and \textit{level-2} subdivides these sections into 88 individual industry classes (A01-U99).
The empirical dataset contains 48,145 German companies. It is crucial to mention, that some of these companies share the same parent company webpage. During the evaluation, the least populated industry classes were omitted, resulting in 19 industry classes in level-1 granularity and 57 classes of level-2 granularity. 

\subsubsection{Company Webpage Data}
In this study, a total number of 41,912 unique company webpages were identified and then used for the process of webscraping. After the NLP preprocessing, each company observation consists of the features \textit{URL}, \textit{HTTP status}, \textit{text}, \textit{image} and \textit{alt}. 

The webscraper is able to reach 38,514 (91.89\%) of the original webpages with an HTTP status code of 200 and at least one output token.

The feature \textit{text} forms the largest part of the output with on average $\mu_{word}=391$ tokens per webpage and a standard deviation of $\sigma_{word}=710$ tokens.
On company webpages, there is only few visual information with $\mu_{img}=4$ ($\sigma_{img}=6$) for the image class labels and $\mu_{alt}=24$ ($\sigma_{alt}=66$) for the ALT tags.

The image classifier is able to assign labels to 65.31\% of the original webpages. Additionally, 74.48\% of the images contained alt tags in the HTML source.

The \textit{URL} is then used as a foreign key to join the additional company information to the webscraping data. The purpose of this additional company information is to evaluate different steps taken in this study. The final Company2Vec model is fully independent of additional company information and solely needs the company URLs.

\subsection{Pretrained Models}

In addition to the company data, self-trained Word2Vec models are compared to three pretrained German models. The first Word2Vec model was trained by Müller\cite{mueller2015} on a Wikipedia and newspaper corpus. The word embeddings have $S=300$ dimensions.
Secondly, the Word2Vec model by Fares et al.\cite{fares-etal-2017-word} was trained with $S=100$ dimensions on a Wikipedia and Gigaword text corpus.
And finally, Yamada et al.\cite{Wikipedia2Vec2018} propose the most recent German Word2Vec model with $S=300$ dimensions. It is fully trained on a Wikipedia corpus. 
All pretrained Word2Vec models did neither apply stemming nor lemmatisation to the training corpus.\\

In the following section, the Word2Vec transformation are applied to the presented data set. The resulting company embeddings are evaluated in terms of industry prediction accuracy. Then, agglomerative clustering, k-means and DBSCAN are applied to the most accurate embeddings for an alternative industry segmentation. Finally, the algorithms for efficient (1) \textit{firm-centric}, (2) \textit{industry-centric} and (3) \textit{portfolio-centric} peer firm identification are explained.

\section{Methodology and Results}\label{Sec:meth_results}
The analytical process follows three steps: First, a machine learning pipeline of different models has been developed, applied and evaluated to create efficient German company embeddings (4.1). Then, a detailed semantic analysis of the resulting company embeddings and industry structures will be presented (4.2). Finally, an accurate and efficient model for peer-firm identification will be discussed (4.3).

Each part of this study will be motivated and described in the following subsections.

\subsection{German Company Embeddings}

\subsubsection{Data Collection}
The elementary input data to the Company2Vec pipeline is an \textbf{URL dataset}, which provides the necessary data for collecting the raw texts and images from companies' webpages. The process of \textbf{webscraping} is performed automatically on \textit{Google Cloud Platform} with a Python script.

Each webpage is opened and searched for visible \textbf{texts}, which are then stored with \textit{latin-1} encoding in a dataframe. Additionally, the \textbf{images} on the webpage are directly classified with a pretrained image classifier and hence transformed into textual data. For this task, the \textit{InceptionResNetV2} classifier with a top-1 accuracy of 0.803 has been selected from Tensorflow. Due to legal regulations, the images are opened temporarily and classified directly without persistent saving of the original data. Only the resulting top-3 image classes are stored in the dataframe. In order to later combine these class labels with the German text, the English class labels are translated into German.
The HTML-alt tags of the images are stored separately.

\subsubsection{Data Preprocessing}

\begin{figure}
    \centering
    \footnotesize
\tikzset{every picture/.style={line width=0.75pt}} 
\begin{tikzpicture}[x=0.75pt,y=0.75pt,yscale=-1,xscale=1]

\draw   (21,41) -- (240.43,41) -- (240.43,74.2) -- (21,74.2) -- cycle ;
\draw   (291,40) -- (480.8,40) -- (480.8,73.2) -- (291,73.2) -- cycle ;
\draw  [fill={rgb, 255:red, 0; green, 0; blue, 0 }  ,fill opacity=1 ] (254,51.61) -- (268.08,51.61) -- (268.08,46) -- (278,57.21) -- (268.08,68.43) -- (268.08,62.82) -- (254,62.82) -- cycle ;

\draw (338,50) node [anchor=north west][inner sep=0.75pt]   [align=left] {``Littfasssaeule"};
\draw (50,50) node [anchor=north west][inner sep=0.75pt]   [align=left] {``\textless/p\textgreater Littfaßsäule \textless/p\textgreater"};
\end{tikzpicture}

\caption{German text example before and after NLP preprocessing.}
\label{Fig:exampleSentence}
\end{figure}
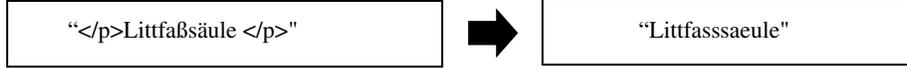

Modern NLP models use specific input formats, which require several preprocessing steps of the raw text from the webpages. Initially, all German special characters are replaced with basic letters and non-alphabetic characters are removed. Existing HTML tags and hyperlinks are removed as well and the resulting string is tokenized. Typical German stopwords and the most frequently used words are excluded, using a dictionary and a counting function. All word tokens with a length of fewer than two characters are removed. An example of this process is presented in figure \ref{Fig:exampleSentence}.

\normalsize

Stemming or lemmatization are not applied, because pretrained language models have a high probability to fail to recognize the shortened word tokens in their vocabulary.

\subsubsection{Data Transformation}
A large body of research literature explores topic modeling on textual data for various applications, e.g. Angelov \cite{Top2Vec}. In this context, a topic model transforms the word tokens into a vectorized representation of a company. The aim is to identify business activities and firms' characteristics in the textual data and to transform them into numeric structures. 

Formally, the following terms are defined according to Blei et al. \cite{BleiLDA}:
\begin{itemize}
    \item A \textit{word} $w_i \in V$, $i \in \{1,..., J\}$ is a basic unit of discrete data, where $V$ is denoted as the \textit{vocabulary} containing all distinct words $V=(w_1, ... , w_{J})$. Note that $J \in \mathbb{N}$ and $J = |V|$.
    \item Each \textit{document} $d_m$, $m \in \{1,..., M\}$ of length $N$ contains a sequence of words $d_m = (w_1, ..., w_{N})$.
    \item A \textit{corpus} is a collection of $M$ documents with $D = (d_1, ..., d_M)$.
\end{itemize}

\paragraph{Word2Vec Topic Modeling.} \label{Meth:W2V}
Efficient semantic vector representations of single words are made possible by using a feed-forward neural network architecture presented by Mikolov et al. \cite{mikolov2013efficient}. Word2Vec is able to model semantic similarities between words based on the context in which they appear in the training data. After training, each word $w_1,..., w_N$  is represented by a sequence of real numbers. Each of these column vectors $\vec{w_i}$ has the same length $S$, called the embedding size. 
The $m$-th document is thus initially transformed into a matrix $\Psi_{m}$ with

\[\Psi_{m} := \left[\begin{array}{c c c}
    \vec{w_{1}} & ... & \vec{w_{N}} \\
    \end{array} \right]_{S\times N}
    \tag{2.1} \label{eq:w2v-mat}
    \]
    

for combining all word vectors of this document.
Mikolov et al. \cite{mikolov2013efficient} already suggest that basic vector operations over multiple word embeddings can be applied depending on the task. For generating one combined document vector, in this study the word embeddings are hence averaged with
\[
\vec{d_m}=N^{-1}\cdot\Psi_m\cdot\mathds{1}_{N\times 1}
\tag{2.2} \label{eq:w2v-mean}
\]
such that the resulting document vector $\vec{d_m}$ is of dimensionality $S\times 1$.

\paragraph{Word2Vec Evaluation.} A self-trained Word2Vec model based on the company webpage corpus is used as a starting point. In the next step, it is benchmarked with pretrained German Word2Vec models. These pretrained models are built on a Wikipedia corpus and/or large newspaper corpora.
The evaluation of word embeddings has been an interest of research for many years: The relatedness between vectorized words is measured with the cosine similarity. These similarity scores are then compared with existing datasets from linguistic studies that empirically quantify word similarities through questionnaires. Two widely used datasets for this evaluation task are \textit{WordSim-353} by Finkelstein et al. \cite{WordSim353:Finkelstein} and \textit{SimLex-999} by Hill et al. \cite{HillRK14}. Both datasets have also been translated into German and will be used for evaluating the quality of the different Word2Vec models.

\subsubsection{Company Embedding Creation}\label{Subsec:companyEmbeddings}
To generate the company embeddings with the above-mentioned topic model (averaging method), different combinations of model input and output are compared.
The preprocessed data consists of (1) word tokens from visible texts, (2) the classified tokens from the images and (3) the HTML-alt tokens. Multiple combinations of these three lists of tokens (1-3) are considered in this process.  Subsequently, a \textit{feature} describes the input list of the preprocessed tokens and a {\small $\overrightarrow{feature}$} is denoted as the vectorized output after transformation:

The following combinations have been created and evaluated:

\newcommand{\svdots}{\raisebox{}{$\scalebox{.75}{\vdots}$}}
\begin{itemize}
    \item Tokens from visible texts only ({\small $\overrightarrow{text}$})
    \item Tokens from image classes only ({\small $\overrightarrow{image}$})
    \item Tokens from alt tags only ({\small $\overrightarrow{alt}$})
    \item Combined by appending the three lists of tokens ({\small $\overrightarrow{text\mathbin\Vert image\mathbin\Vert alt}$})
    \item Combined by concatenating all three individual vectors ({\small $\overrightarrow{text}\mathbin\Vert \overrightarrow{image}\mathbin\Vert \overrightarrow{alt}$})
\end{itemize}

While training, the above-mentioned language models depend on large text corpora in which the order of words might be crucial. For that reason, all self-trained language models are trained on the visible \textit{text} tokens only. After training, these models can infer topics/vectors from the unseen data columns \textit{image} and \textit{alt}.

\paragraph{Evaluation.}
Company2Vec was introduced as an unsupervised learning technique. However, this condition is softened in order to evaluate the different unsupervised embedding techniques and select the most accurate company embeddings. 
Considering companies' official industry sectors, a proxy for companies' similarities is already given, because similar companies tend to share the same industry sector. This information is hence used as a target feature for a multiclass classification as proposed by Pierre \cite{Pierre2001OnTA} and Berardi et al. \cite{BerardiEtAl2015}. 

This multiclass classification compares accuracy-scores of different combinations in above mentioned company embedding techniques. Four classifiers are tested in this evaluation process: A \textit{logistic regression}, a \textit{random forest}, a \textit{k-nearest neighbors} (kNN) classifier and a \textit{neural network}. Due to the amount of different token combinations, embedding methods, classification algorithms and hence computational limitations, a less time-consuming \textit{train-test-split} is preferred over a full k-fold cross validation.

Furthermore, potential problems of class imbalance have already become apparent in Bernstein et al. \cite{BernsteinEtAl2003}. However, the authors do not consider these problems further. This study at hand solves the class imbalance issues with \textit{Synthetic Minority Over-sampling Technique} (SMOTE) of minority classes and \textit{random undersampling} of majority classes. \cite{Chawla_2002} This treatment is applied to the training data after performing the train-test-split. Thus, the models are trained on balanced data and the test set remains unbalanced for unbiased evaluation.

\paragraph{Principal Component Analysis}
After data transformation and after choosing the best company embedding method in terms of industry prediction accuracy, the pre-final company vectors are analysed for correlations. 
A \textit{principal component analysis} (PCA) is used as an unsupervised technique for dimensionality reduction. With a PCA, the embedding dimensionality $S$ can be efficiently reduced to $S'<S$ while sustaining the ``explained variance" $\mathbb{V}_S$ of the original data. 

In order to find a suitable $S'$, the amount of variance reduction is analysed under the condition that $\mathbb{V}_{S'}$ must explain over 90\% of the original variation. 

\paragraph{Visualisation}
As an additional indicator for the distribution of company embeddings, several visualisation techniques are applied. One basic visualisation of the classification results are heatmaps of confusion matrices. Furthermore, the final company embeddings are plotted using a \textit{t-Distributed Stochastic Neighbor Embedding} (t-SNE), which reduces the dimensionality of the company embeddings to $S''=2$. \cite{vanDerMaaten2008}

The aim is to visualize relations between German companies on a 2-dimensional map for gaining insights into German company and industry structures.

\subsubsection{Results} \label{Sec:tranformation}

In contrast to the other language models (such as LDA or Doc2Vec), Word2Vec company embeddings require a transformation of each word token into word embeddings with the trained Word2Vec model. This additional step is performed both on the pretrained and self-trained models. After transformation, multiple company embedding strategies are assessed with the true industry classes and the best method will be highlighted. 

\paragraph{Word Embedding Evaluation.}
The idea is to transform the word tokens with the most accurate embeddings in terms of true semantic relatedness between words (see section \ref{Meth:W2V}). Five Word2Vec models are assessed on WordSim-353 and SimLex-999 using the correlation between empirical word similarities (ground truth) and the models' cosine similarities. The results (see appendix, table \ref{tab:word2vecs}) show that self-trained models on the company webpage corpus do not compete with publicly available pretrained models.
For instance, comparing the correlations of the self-trained model B (0.0609) with the correlation of Müller \cite{mueller2015} (0.5861) shows a large difference.
The evaluation of self-trained models identifies a negligible correlation between true word-pair relations and the Word2Vec similarity scores. Furthermore, only 81.59\% (WordSim-353) respectively 77.78\% (SimLex-999) of the words in the two evaluation sets appear in the training corpus. Independent from model specifications, the presented training corpus is hence too small to achieve competitive results.

The pretrained models perform significantly better in this evaluation: With a low/moderate positive correlation, all three pretrained models are able to replicate semantic word relationships. With a \textit{vocabulary coverage} of over 97\%, Müller \cite{mueller2015} identifies the most word-pairs of the evaluation sets. This model also performs best on SimLex-999 with a low positive correlation of 0.3678 and offers a moderate positive correlation on WordSim-353. The storage format of the presented model does not allow for training additional epochs. Therefore, Müller \cite{mueller2015} is selected as the final model for this data transformation.

\paragraph{Company Embedding Evaluation.} \label{Par:compEmbEval}
The Word2Vec approach offers multiple combinations and techniques for generating company embeddings, which vary in their predictive power (see section \ref{Subsec:companyEmbeddings}). Table \ref{tab:word2vec19} forms the first part of the assessment with 19 distinct industry classes (\textit{level-1}). In order to overcome problems concerning class imbalance, the results are validated with a balanced training set of the company data using SMOTE.

Comparing the classification results for the original features \textit{text}, \textit{image} and \textit{alt}, table \ref{tab:word2vec19} shows the significance of textual data on webpages. Using the original \textbf{imbalanced data} in the \textit{level-1} prediction task, the logistic regression on {\small $\overrightarrow{text}$} with $S=300$ dimensions predicts over $66\%$ of the 19 industry classes correctly. Compared to the na\"ive solution of always selecting the most frequent class $Y_{19}=$'C' with a baseline probability of $P_{19}(Y=\ $'C'\ $|\ X)=P_{19}(Y=\ $'C'$)=0.2948$, the variable {\small $\overrightarrow{text}$} shows strong predictive power.


\bgroup
\def\arraystretch{1.2}
\begin{table} \centering 
\footnotesize
  \caption{Classification results of multiple embedding strategies with 19 distinct industries.} 
  \label{tab:word2vec19} 
\begin{tabular}{@{\extracolsep{13pt}} lcccccc} 
\toprule
& Embedding & Top-N & LogR & kNN & RF & NN\\
\midrule
\parbox[t]{2mm}{\multirow{10}{*}{\rotatebox[origin=c]{90}{IMBALANCED*}}}& \multirow{2}{*}{$\overrightarrow{text}$} &top-1 & \textbf{0.6606} & 0.6358 & \textbf{0.5894} & \textbf{0.6737}\\
\vspace{5px}
 & &  top-3  & \textbf{0.9008} &  0.8295 & 0.8150 & \textbf{0.9011}\\
 & \multirow{2}{*}{$\overrightarrow{image}$}  &top-1 & 0.3938 & 0.3804 & 0.3979 & 0.3888\\
 \vspace{5px}
 &  &top-3 & 0.7291 &  0.6251 & 0.6732 & 0.6920\\
 & \multirow{2}{*}{$\overrightarrow{alt}$}  & top-1 & 0.4845 & 0.4691 & 0.4582 & 0.4852\\
 \vspace{5px}
 & & top-3 & 0.7887 &  0.6953 & 0.7055 & 0.7501\\
 & \multirow{2}{*}{$\overrightarrow{text\mathbin\Vert image\mathbin\Vert alt}$} & top-1 & 0.6577 & \textbf{0.6398} & 0.5838 & 0.6635\\
 \vspace{5px}
 &  &top-3 & \textbf{0.9008} &   \textbf{0.8314} & \textbf{0.8154} & 0.8966\\
 & \multirow{2}{*}{$\overrightarrow{text}\mathbin\Vert \overrightarrow{image}\mathbin\Vert \overrightarrow{alt}$} & top-1 & 0.6559 & 0.4611 & 0.5630 & 0.5906\\
  \vspace{2px}
 &  & top-3 & 0.8915 &  0.6991 & 0.8050 & 0.8307\\
\multicolumn{7}{l}{\scriptsize{* Best na\"ive top-1 solution: 0.2948 (predicting always class C)}}\\

\midrule
\parbox[t]{2mm}{\multirow{10}{*}{\rotatebox[origin=c]{90}{BALANCED**}}}&  \multirow{2}{*}{$\overrightarrow{text}$} &top-1 & \textbf{0.5677} &  0.4404 & 0.4744 & \textbf{0.5712} \\
 \vspace{5px}
 &  &  top-3  & \textbf{0.8148} &   0.6484 & \textbf{0.6608} & \textbf{0.8333}\\
 &  \multirow{2}{*}{$\overrightarrow{image}$}  &top-1 & 0.2004  & 0.1690 & 0.2118 & 0.2324\\
  \vspace{5px}
 & &top-3 & 0.4367 &  0.3569 & 0.3687 & 0.4990 \\
 &  \multirow{2}{*}{$\overrightarrow{alt}$}  & top-1 & 0.3354 & 0.2467 & 0.2783 & 0.3760 \\
  \vspace{5px}
 & & top-3 & 0.5905 &  0.4472 & 0.4362 & 0.6489 \\
 & \multirow{2}{*}{$\overrightarrow{text\mathbin\Vert image\mathbin\Vert alt}$} & top-1 & 0.5630 & \textbf{0.4451} & \textbf{0.4786} & 0.5613 \\
   \vspace{5px}
 &   &top-3 & 0.8100 &  \textbf{0.6619} & \textbf{0.6608} & 0.8294 \\
 & \multirow{2}{*}{$\overrightarrow{text}\mathbin\Vert \overrightarrow{image}\mathbin\Vert \overrightarrow{alt}$} & top-1 & 0.5325 & 0.2285 & 0.4442 & 0.4976 \\
 \vspace{2px}
 & & top-3 & 0.7913 &  0.4412 & 0.6243 & 0.7662 \\
 \multicolumn{7}{l}{\scriptsize{** A priori top-1 solution: 0.0526 (random selection)}}\\

\bottomrule
\end{tabular} 
\end{table} 
\normalsize

One restricting factor of a classification model lies within the fact that the industry membership of a company is limited to one distinct sector. This study includes the additional top-3 predictions in the evaluation. Using top-3 predictions, the model classifies over $90\%$ of the companies correctly. The neural network is able to slightly improve these scores. On the one hand, textual data has the most impact on the predictive accuracy. On the other hand, the visual data such as \textit{image} or \textit{alt} is still able to outperform the na\"ive solution.

Another promising explanatory feature on 19 industry classes is {\small $\overrightarrow{text\mathbin\Vert image\mathbin\Vert alt}$}, where the textual and visual information is combined in a company vector of dimensionality $S=300$. The accuracy is almost identical to the textual data. Here, the \textit{kNN} model and the \textit{random forest} can improve their accuracy scores even further.

The four selected models are trained on \textbf{balanced data} and tested on original (imbalanced) hold-out data. These steps are important to learn class-specific features and to validate the results. Considering balanced data, the a priori probability is reduced to $P_{19}(Y)=0.0526$ under random selection. This baseline probability is outperformed by a top-1 accuracy of $0.5677$ of the logistic regression.
The data shows that the models are able to classify the majority of companies correctly after treating the industry class imbalance.


As mentioned above, this evaluation aims to generate a company embedding technique that retains as much information about the company's business activities as possible. Table \ref{tab:word2vec57} illustrates the classification results of a \textit{level-2} granularity (57 industries) for the same models and embeddings. The model Company2Vec, developed in this study, should have the ability to measure fine-granular distances between companies.

Clearly, the accuracy scores decrease due to predicting three times as many industry sectors as the \textit{level-1} granularity classification. Among the \textbf{imbalanced data}, the baseline probability decreased to $P_{57}(Y=\ $'G46'\ $|\ X)=P_{57}(Y=\ $'G46'$)=0.1831$ for a na\"ive classifier predicting always the most frequent class $Y_{57}=$'G46'. Again, the company embeddings outperform this baseline when combining the textual and visual features. On a more fine-granular industry segmentation level, the company embeddings benefit from the additional information provided by the visual data. The logistic regression on {\small $\overrightarrow{text\mathbin\Vert image\mathbin\Vert alt}$} predicts $46.44\%$ of the classes directly and $71.25\%$ of the classes using the top-3 predictions. The neural network improves this accuracy further to $50.82\%$ (top-1) and $74.75\%$ (top-3).  

After treating the class imbalance, the discussed embedding strategy still provides the best results. Considering random selection, the a priori solution has decreased to $P_{57}(Y)=0.0167$. The evaluation on the (imbalanced) hold-out testing data states that the classifiers identify class-specific characteristics once more using the company embedding with {\small $\overrightarrow{text\mathbin\Vert image\mathbin\Vert alt}$}. A visualisation of the confusion matrix is given in figure \ref{Fig:conf_level2-w2v} (see appendix).

In summary, using the Word2Vec approach, a combination of all three features \textit{text}, \textit{image}, \textit{alt} provides the most accurate company embeddings in fine-granular industry classification systems.

\bgroup
\def\arraystretch{1.2}
\begin{table}[H] \centering 
\footnotesize
  \caption{Classification results of multiple embedding strategies with 57 distinct industries.} 
  \label{tab:word2vec57} 
\begin{tabular}{@{\extracolsep{13pt}} lcccccc} 
\toprule
& Embedding & Top-N & LogR & kNN & RF & NN\\
\midrule
\parbox[t]{2mm}{\multirow{10}{*}{\rotatebox[origin=c]{90}{IMBALANCED*}}}& \multirow{2}{*}{$\overrightarrow{text}$} &top-1 &  0.4578 & 0.4200 & 0.3796 & 0.5019 \\
\vspace{5px}
 & &  top-3  & 0.7082 &  0.6212 & \textbf{0.5519} & 0.7382 \\
 & \multirow{2}{*}{$\overrightarrow{image}$}  &top-1 & 0.2528 & 0.1967 & 0.2410 & 0.2464 \\
 \vspace{5px}
 &  &top-3 & 0.4584 &   0.3669 & 0.3613 & 0.4283 \\
 & \multirow{2}{*}{$\overrightarrow{alt}$}  & top-1 & 0.3162 & 0.2756 & 0.2773 & 0.3237 \\
 \vspace{5px}
 & & top-3  & 0.5381 &   0.4494 & 0.3952 & 0.5026 \\
 & \multirow{2}{*}{$\overrightarrow{text\mathbin\Vert image\mathbin\Vert alt}$} & top-1 & \textbf{0.4644}  & \textbf{0.4271} & \textbf{0.3868} & \textbf{0.5082} \\
 \vspace{5px}
 &  &top-3  & \textbf{0.7125} &  \textbf{0.6265} & 0.5421 & \textbf{0.7475} \\
 & \multirow{2}{*}{$\overrightarrow{text}\mathbin\Vert \overrightarrow{image}\mathbin\Vert \overrightarrow{alt}$} & top-1 & 0.4577  & 0.2632 & 0.3651 & 0.4066 \\
  \vspace{2px}
 &  & top-3 & 0.6970 &   0.4482 & 0.5206 & 0.6217 \\
\multicolumn{7}{l}{\scriptsize{* Best na\"ive top-1 solution: 0.1831 (predicting always class G46)}}\\

\midrule

\parbox[t]{2mm}{\multirow{10}{*}{\rotatebox[origin=c]{90}{BALANCED**}}}&  \multirow{2}{*}{$\overrightarrow{text}$} &top-1 & 0.3818  & 0.28100 & 0.2794 & 0.3952 \\
 \vspace{5px}
 &  &  top-3  & 0.5904 &   0.4489 & \textbf{0.3993} & 0.6290 \\
 &  \multirow{2}{*}{$\overrightarrow{image}$}  &top-1 & 0.1487  & 0.1061 & 0.1303 & 0.1433 \\
  \vspace{5px}
 & &top-3  & 0.2590 &  0.2254 & 0.2094 & 0.2716 \\
 &  \multirow{2}{*}{$\overrightarrow{alt}$}  & top-1 & 0.2348 & 0.1563 & 0.1632 & 0.2317 \\
  \vspace{5px}
 & & top-3 & 0.3686 &  0.2954 & 0.2416 & 0.3864 \\
 & \multirow{2}{*}{$\overrightarrow{text\mathbin\Vert image\mathbin\Vert alt}$} & top-1  & \textbf{0.3886} & \textbf{0.2842} & \textbf{0.2775} & \textbf{0.4086} \\
   \vspace{5px}
 &   &top-3 & \textbf{0.5974} &  \textbf{0.4557} & 0.3930 & \textbf{0.6462} \\
 & \multirow{2}{*}{$\overrightarrow{text}\mathbin\Vert \overrightarrow{image}\mathbin\Vert \overrightarrow{alt}$} & top-1 & 0.3522 & 0.1476 & 0.2580 & 0.3250 \\
 \vspace{2px}
 & & top-3 & 0.5569 &   0.2867 & 0.3702 & 0.5243 \\
 \multicolumn{7}{l}{\scriptsize{** A priori top-1 solution: 0.0167 (random selection)}}\\

\bottomrule
\end{tabular} 
\end{table} 
\normalsize

\paragraph{Model Selection.}
The evaluation results of various company embedding techniques expresses the power of pretrained word embeddings and the importance of combining textual and visual data of company webpages. For fine-granular industry representation, a company embedding with {\small $\overrightarrow{text\mathbin\Vert image\mathbin\Vert alt}$} outperforms other embedding techniques in terms of accuracy. This embedding robustly deals with class imbalances and provides sensible information about the business activities of a company.

\paragraph{Dimensionality Reduction.} \label{Sec:DimRed}

The proposed embedding technique produces company vectors with $S=300$ dimensions. Analysing the correlation of the company vectors, a positive relatedness can be uncovered (Fig. \ref{Fig:PCA-normalisation}).

\begin{figure}
\centering
\includegraphics[trim=15 15 50 30, clip,width=0.49\textwidth]{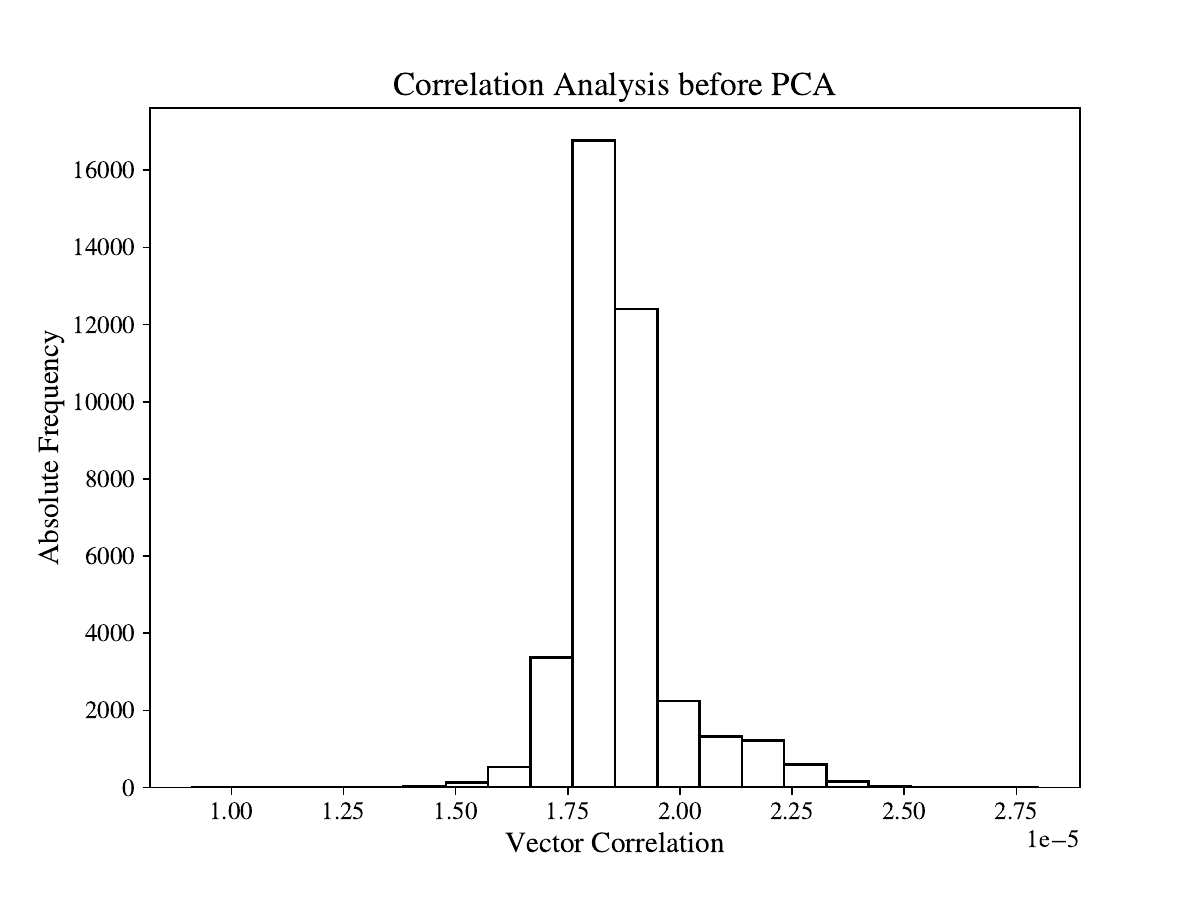}
\includegraphics[trim=15 15 57 30, clip,width=0.49\textwidth]{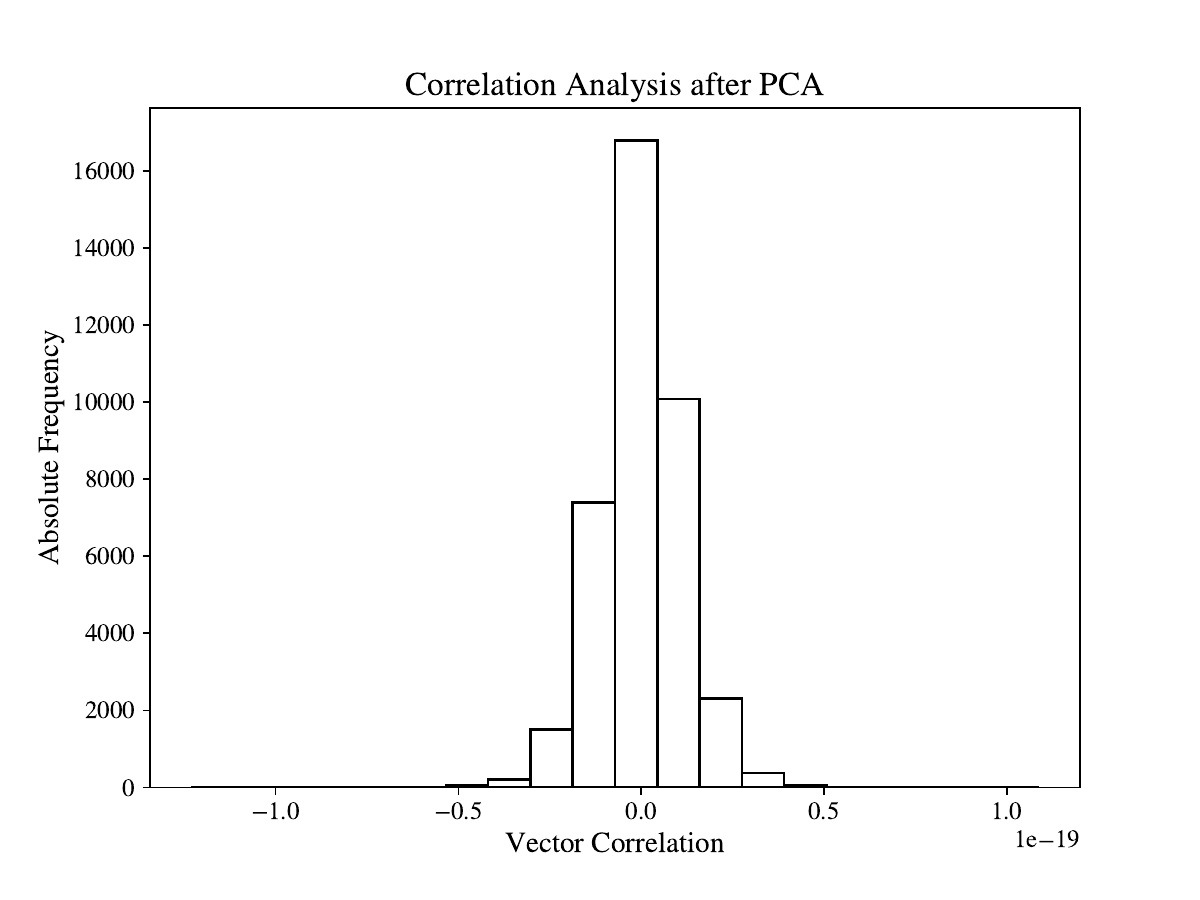}
\caption{Vector correlation before (left) and after (right) PCA.}
\label{Fig:PCA-normalisation}
\end{figure}

The correlation might originate from the fact that all vectors represent some kind of company, which are similar real-world objects. Mathematically, the vector averaging method produces similar average values on certain dimensions. Consequently, those dimensions which are similar among a majority of firms can be omitted using \textit{dimensionality reduction}.   

A principal component analysis treats the positive vector correlation and transforms the dimension of each vector to a smaller dimension. As final embedding size, a dimensionality of $S'=100$ is chosen, where about 94\% (see appendix, Fig. \ref{Fig:pca}) of explained variance is maintained.

The impact of this transformation is demonstrated in table \ref{tab:pca}:
The cosine similarity before and after PCA is presented for multiple company pairs. The dimensionality transformation normalizes company similarities and thus enhances the interpretability of companies' relations. Likewise, two companies with similar business activities such as \textit{Warner Music Group} and \textit{Universal Music} maintain a high similarity score of $0.9278$. In contrast, two rather different companies such as \textit{Commerzbank} and \textit{McDonald's Restaurants}, which originally shared a high positive similarity score of $0.7848$, now have a negative score ($-0.1539$) after the transformation.

In summary, the dimensionality reduction (PCA) to $S'=100$ dimensions is an important part of the Company2Vec model and improves the model's interpretability. Reducing the dimensionality of the embeddings, run-time and memory complexity are lowered substantially for the subsequent peer-firm identification process.

\bgroup
\def\arraystretch{1.2}
\begin{table}[H] \centering 
\footnotesize
  \caption{Refinement of company similarities using a PCA.} 
  \label{tab:pca} 
\begin{tabular}{@{\extracolsep{15pt}} llcc} 
\toprule
\multirow{2}{*}{Company 1} & \multirow{2}{*}{Company 2} & \multicolumn{2}{c}{ Cosine Similarity}\\
\cline{3-4}

& & before PCA & after PCA\\
\midrule
Airbus & Rheinmetall  & 0.9759 & 0.9519 \\
Apple  & LG Electronics   & 0.9721 & 0.9519 \\
Universal Music & Warner Music Group  & 0.9674 & 0.9278 \\
Commerzbank & Deutsche Kreditbank  & 0.9821 & 0.9167\\
GameStop & Nintendo & 0.9493 & 0.8232\\
Hugo Boss & Lacoste & 0.9446 & 0.8158\\
Dallmayr \textit{(coffee brand)} & Melitta \textit{(coffee brand)} & 0.9310 &0.7963\\
LEGO & Nintendo & 0.9400 & 0.7948\\
EDEKA \textit{(grocery)} & McDonald's Restaurants & 0.8633 & 0.3018 \\
Flixbus \textit{(traveling service)} & Microsoft & 0.7701 & -0.0464\\
Deutsche Kreditbank & Hugo Boss & 0.7692 & -0.1303 \\
Commerzbank & McDonald's Restaurants  & 0.7848 & -0.1539\\
Apple & Cornelsen \textit{(publisher)} & 0.6286 & -0.4845\\
Bauhaus \textit{(hardware store)} & ROLEX & 0.6229 & -0.5455\\
Rheinmetall & Cornelsen \textit{(publisher)} & 0.5092 & -0.6877 \\
\bottomrule
\end{tabular} 
\end{table} 
\normalsize


\subsection{Semantics of Company Embeddings} \label{Sec:Semantics}
The positioning of word embeddings in a vector space represents semantic relations between words. Word2Vec automatically captures semantic concepts and implicitly learns word relationships without supervised information. \cite{DBLP:journals/corr/MikolovSCCD13} This subsection explores semantic patterns of the final Company2Vec embeddings. By analysing \textit{top-n words} for companies, semantic \textit{company-word relations} and \textit{industry structures}, the semantic relationship between company embeddings is assessed.



\subsubsection{Top-N Words}

\bgroup
\def\arraystretch{1.1}
\begin{table}[H] \centering 
\footnotesize
  \caption{Top-N words for \textit{Deutsche Wohnen}, \textit{Commerzbank} and \textit{1\&1}.} 
  \label{tab:top-n} 
\begin{tabular}{@{\extracolsep{30pt}} lll} 
\toprule
Deutsche Wohnen  & Commerzbank & 1\&1 \\

\textit{(real estate/property)} & \textit{(commercial banking)} & \textit{(telecommunications)} \\
\midrule
Wohnen          &  Girokonto &  Mobilfunknetz \\
\textit{(living)}    &  \textit{(bank account)} &  \textit{(mobile network)} \\\rule{0pt}{3ex}
bezahlbaren\_Wohnraum   & abbuchen  &  Mobilfunk \\
\textit{(affordable housing)}    &  \textit{(withdraw)} &  \textit{(cellular)} \\\rule{0pt}{3ex}
Wohneigentum	        & Guthaben  & Telefonie  \\
\textit{(home ownership)}    &  \textit{(deposits)} &  \textit{(telephony)} \\\rule{0pt}{3ex}
sozialen\_Wohnungsbaum   &  per\_Lastschrift & Breitband-Internetzugang  \\
\textit{(social housing)}    &  \textit{(via debit charge)} &  \textit{(broadband networks)} \\\rule{0pt}{3ex}
Mieten	                & Kreditnehmer  &  Datennetz \\
\textit{(renting)}    &  \textit{(borrower)} &  \textit{(data network)} \\\rule{0pt}{3ex}
Wohnraum	            & Sparkonto  & VDSL\\
\textit{(living space)}    &  \textit{(savings account)} &  \textit{(Very high-speed DSL)} \\

\bottomrule
\end{tabular} 
\end{table} 
\normalsize

Since Company2Vec derives from topic modeling approaches, the embedding holds information about the business activities of a company. In order to get this information, the most similar words for a company embedding are determined by the presented Word2Vec model. Beforehand, the company-based PCA transformer (\ref{Subsec:companyEmbeddings} and \ref{Sec:DimRed}) has been applied to the word embeddings for vector space harmonisation.  

The entries in table \ref{tab:top-n} confirm the hypothesis that Company2Vec embeddings are able represent firms' business activities. For instance, the most similar words for the real estate company \textit{Deutsche Wohnen} indicates a business model concerning living and housing. This finding leads to a potential application, which is generating automated textual company descriptions.

\subsubsection{Company-Word Relation} 

\begin{figure}[h]
\centering
\includegraphics[trim=100 60 106 65, clip,width=1\textwidth]{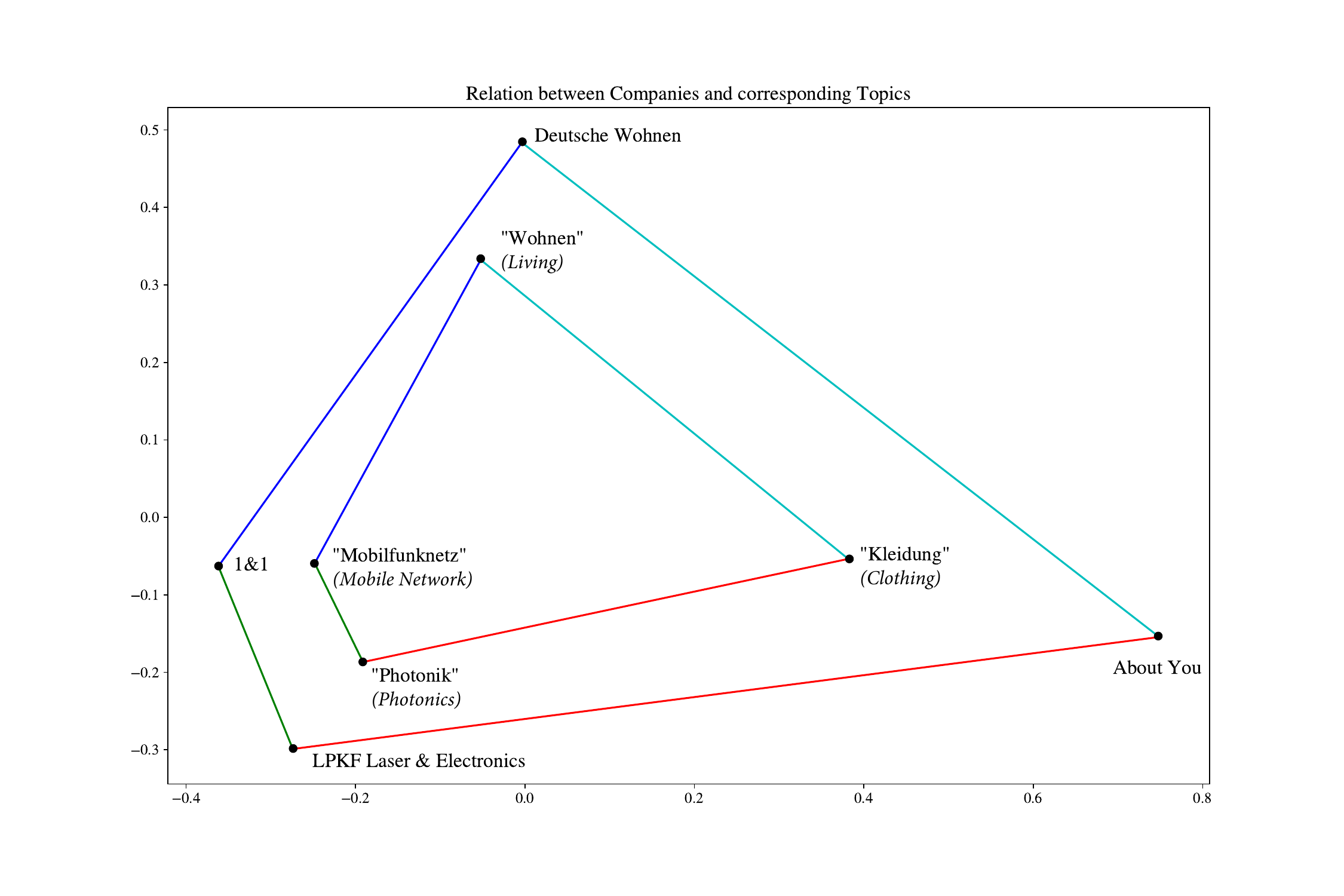}
\caption{Four companies in relation with their most similar word.}
\label{Fig:company_topic_relation_1}
\end{figure}

To further scrutinize the relation between company and word embeddings, figure \ref{Fig:company_topic_relation_1} presents a sample of four companies. For each, the corresponding most-similar word is included. The companies are closely located to the topics that describe their business activities. By adding connecting lines between the topic positions, a shape is formed that represents the semantic relatedness in distance and direction. This topic shape is proportionally almost identical to the company shape. Hence, the company embeddings and word vectors can be integrated in one composite embedding space. This allows further universal applications and enhances interpretability.

In addition, semantic analogies (Fig. \ref{Fig:company_topic_relation_2}) that apply to words can be transferred to companies. A retail bank and a building society both offer financial services. However, the portfolio of a building society (\textit{Wüstenrot Bausparkasse}) is more focussed on real estate loans. The difference of a building society compared to a retail bank (\textit{Commerzbank}) follows the difference between ``\textit{bank account}" and ``\textit{buying a house}".

\begin{figure}[h]
\centering
\includegraphics[trim=100 60 90 65, clip,width=1\textwidth]{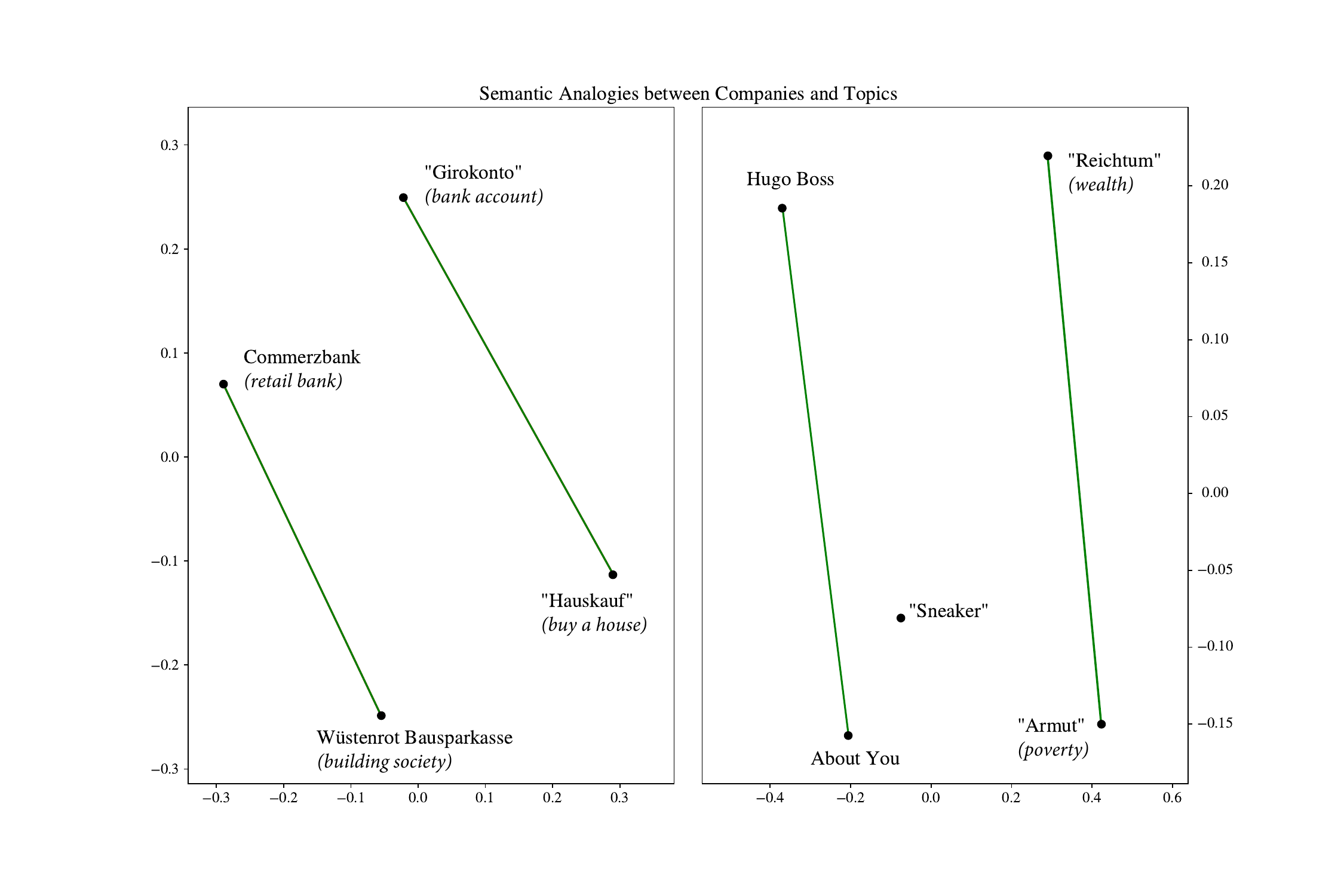}
\caption{Semantic analogies between companies and words.}
\label{Fig:company_topic_relation_2}
\end{figure}

Another example describes the relation between the luxury brand \textit{Hugo Boss} and the online fashion retailer \textit{About You}. The comparison between ``\textit{wealth}" and ``\textit{poverty}" forms an analogy that can be transferred to the relation of both companies. Company2Vec thus distinguishes fine-granularly between companies with similar business models and reflects typical semantic attributes.

\subsubsection{Industry Structure} 
The final iteration of this semantic analysis is a t-SNE visualisation of some German companies (Fig. \ref{Fig:TSNE-examples}). They stem from different industrial sectors and are transformed into a semantic industry map. The distances and hence the coordinates of the firms are derived from the company embeddings. Some areas on the map form groups of companies with similar business models. Grocery vendors (\textit{e.g. EDEKA}) are positioned close to food manufacturers (\textit{e.g. Dr. Oetker}). Hospitals are close to chemical-pharmaceutical companies and care facilities. Banks (\textit{e.g. Commerzbank}) and insurance companies (\textit{e.g. Feuersozietät}) are also grouping together. The first ``semantic averages" become apparent: The car bank (\textit{Mercedes-Benz Bank}) is located between the retail banks and car sellers. And finally, the hardware stores (\textit{e.g. Bauhaus}) is positioned between supermarkets and housing associations (\textit{e.g. Degewo}).

Thus, Company2Vec represents the German industry semantically correct using the proposed language model.

\begin{figure}
\centering
\includegraphics[trim=98 90 114 85, clip,width=1\textwidth]{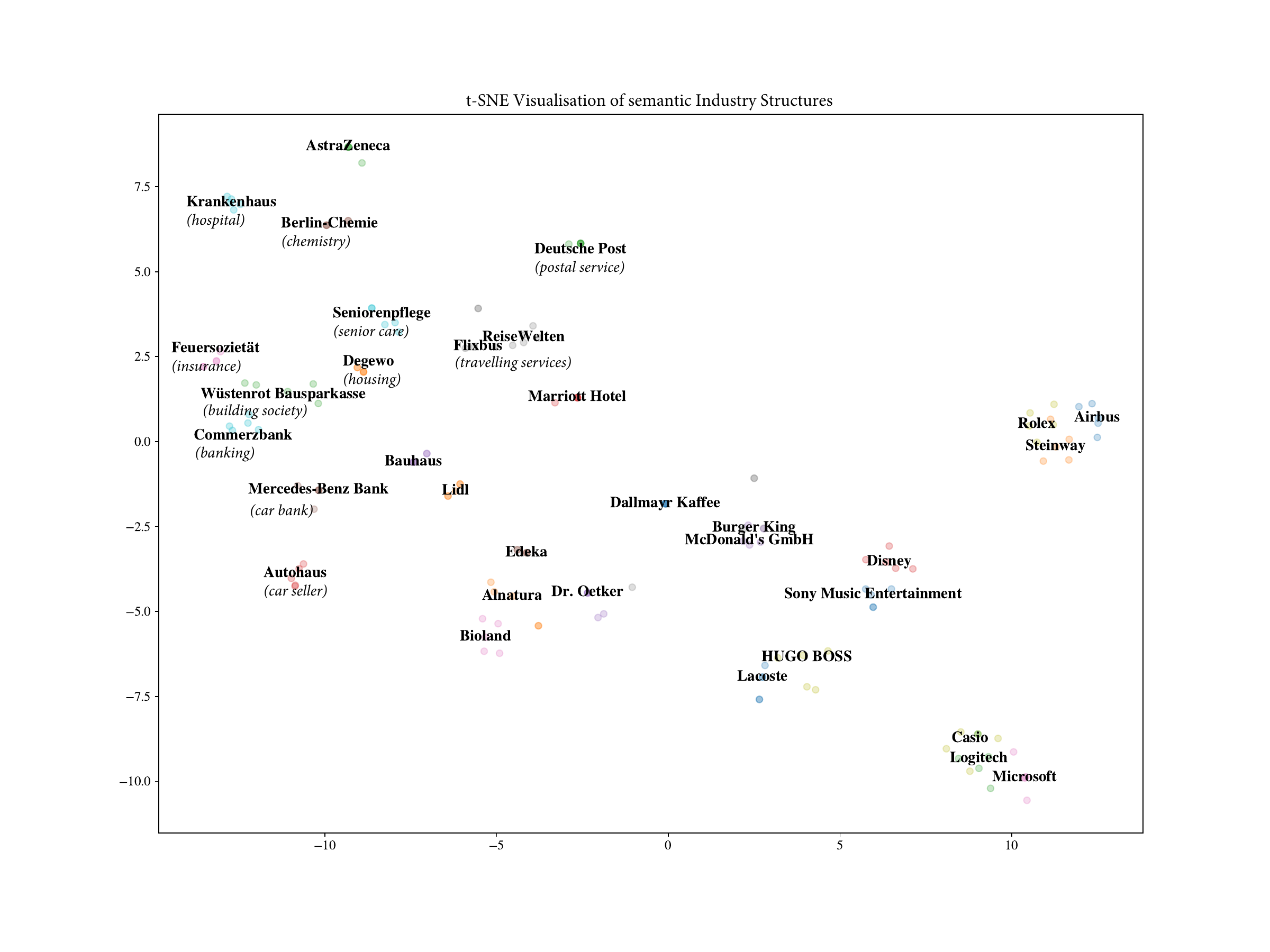}
\caption{Industry structures in a t-SNE visualisation of sampled companies.}
\label{Fig:TSNE-examples}
\end{figure}


\subsection{Peer Firm Identification}
The final company embeddings are able to represent industry structures accurately. Therefore they can be used in fulfilment of the research objective of this work: peer firm identification. Due to missing data, it is not possible to assign company vectors to all firms. In the used model, these firms are represented by an empty vector and are not included in the following task.

\subsubsection{Similarities}
Numeric company embeddings offer a straightforward calculation of firms' similarities. These similarities are necessary for estimating companies' distances in cluster analyses. Two distance measures are considered for this task:

A standard distance measure for numeric data is the \textbf{Euclidean distance} as presented in equation \ref{eq:distEUC}. The geometric interpretation of two companies $f_1$ and $f_2$ is a representation with two data points in a high-dimensional space.

\[
dist(f_1, f_2) = |\vec{f_1} -\vec{f_2}|
\tag{3.1} \label{eq:distEUC}
\]

One advantage of vectorized company embeddings is the possibility to compute the \textbf{cosine distance} alternatively.
\[
dist(f_1, f_2) = 1 - cos(\Theta) = 1-\frac{\vec{f_1} \cdot \vec{f_2}}{|\vec{f_1}| \cdot |\vec{f_2}|}
\tag{3.2} \label{eq:distCOS}
\]

The angle $\Theta$ between two company embeddings $\vec{f_1}$ and $\vec{f_2}$ indicates the divergence between these two vectors. \cite{Singhal2001} The cosine projects the angle to a numeric similarity score. The resulting cosine distance between two firms $f_1$ and $f_2$ is presented in equation \ref{eq:distCOS}.

Both distance measures are used in the following subsection.

\subsubsection{Clustering}
Three different clustering algorithms are used for creating peer firm clusters. \textbf{Agglomerative clustering} offers a hierarchical solution, using the cosine distance for segmentation. Single-linkage, average-linkage and complete linkage will be tested.

Another clustering algorithm that is able to train segments on the cosine distance is \textbf{DBSCAN}. It identifies dense areas and removes outlying noise points. The number of clusters can only be influenced indirectly by density parameters.

The third algorithm, \textbf{k-means}, does not work with cosine distances and is hence trained on the Euclidean distances. Considering k-means, the number of firm segments is initially set before training. With the typical cluster centroids, k-means represents the geometric center of each cluster.

\subsubsection{Evaluation} 

In order to identify a proper clustering of the company embeddings, the clusters are then used in a classification task. This is a necessary step because the clusters of company embeddings should represent industry segments.
In agglomerative (hierarchical) clustering and k-means, the final number of clusters can be selected manually. The optimal number of clusters is assessed by comparing the prediction accuracies. The prediction accuracy indicates the purity of clusters because only homogeneous industry clusters perform well in an industry prediction task. Further quantitative and visual comparisons are applied.

\subsubsection{Recommendation}
Depending on the final Company2Vec pipeline, three algorithms are developed for \textit{firm-centric}, \textit{industry-centric} and \textit{portfolio-centric} peer firm identification. The proposed algorithms are described and assessed in terms of memory and run-time complexity.

Finally, for each of these three algorithms, some examples of German firms are presented. The results give insights into German company similarities and industry structures.  

\subsubsection{Results}

\paragraph{Clustering.}\label{Sec:ClusteringFunction} First, hierarchical (agglomerative) clustering and DBSCAN have been tested under multiple parameter combinations. Both methods generate only few but large clusters. Hence, they do not prove to be suitable for Company2Vec due to lacking fine-granularity. 

In contrast to agglomerative clustering and DBSCAN, a company segmentation with \textbf{k-means} requires a predefined number of clusters $k$. Based on the company data, the algorithm is able to form well-proportioned segments for peer firm identification. Therefore, an optimal number of clusters is assessed in the following steps. 
For visualisation purposes, a lower number of $k=50$ clusters is used to demonstrate the company segment proportioning in figure \ref{Fig:TSNE-kmeans-50} (see appendix).

To identify a suitable number of clusters, the evaluation strategy does not only focus on model-intrinsic scoring methods but covers the task of industry-centric peer firm identification as well. For model-intrinsic assessment, the \textit{elbow method} is used (see appendix, Fig. \ref{Fig:kmeans-elbow}) to analyse the distortion score for each $k$. This method proposes a small number of $15\leq k\leq 25$ clusters.

However, a larger number of clusters is required to ensure fine-granular industry segments. A characteristic of k-means is creating geometric centroids per cluster to represent the center of a segment. These 100-dimensional cluster centroids are used in the same industry prediction setting as the original company embeddings. For a full examination, the prediction accuracies for both level-1 and level-2 industry systems are given in figure \ref{Fig:clust-kmeans-numclust}.

 \begin{figure}[h]
\centering
\includegraphics[trim=100 40 115 40, clip,width=1\textwidth]{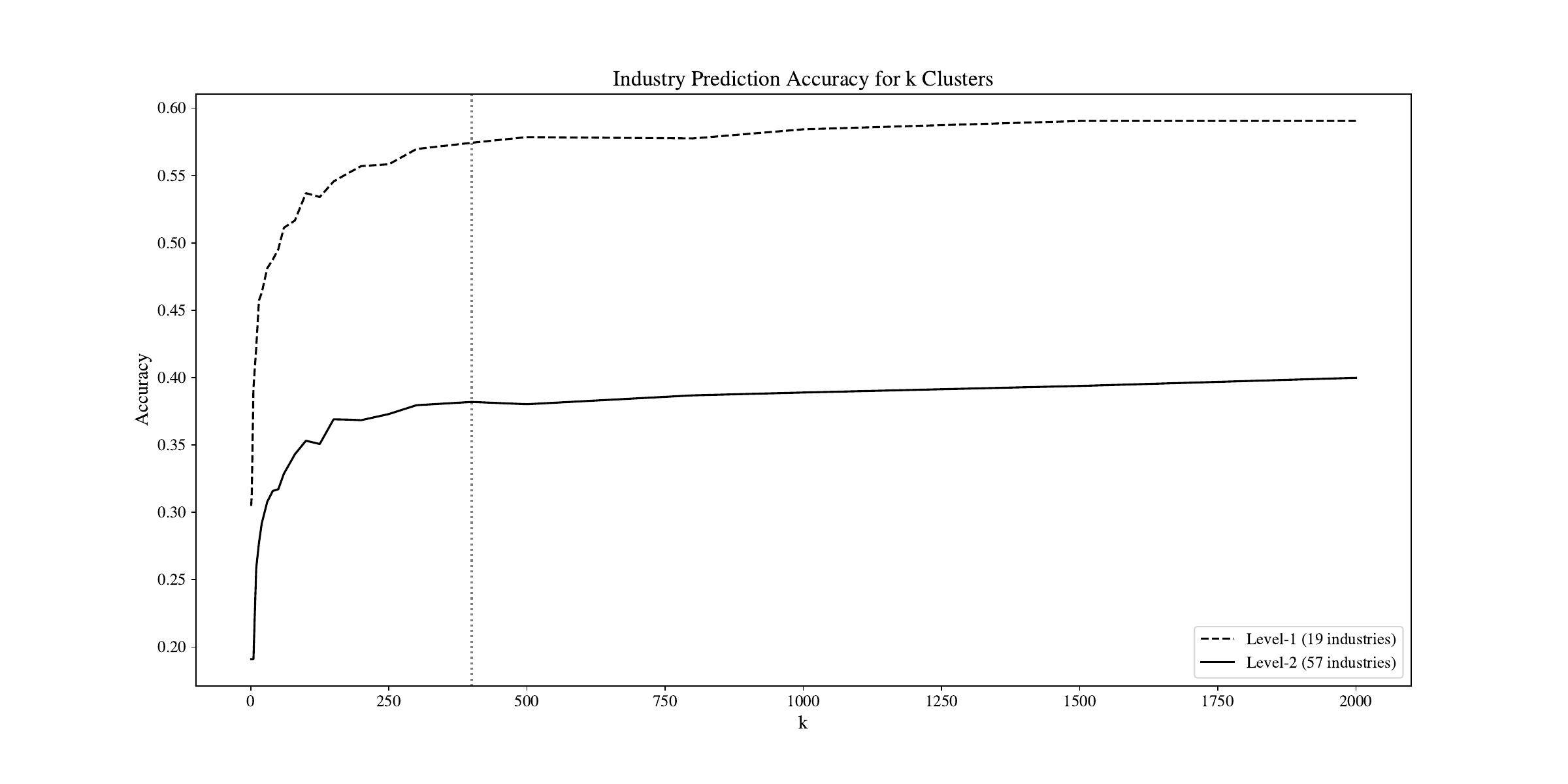}
\caption{Industry prediction accuracy on k-means cluster centroids for k clusters.}
\label{Fig:clust-kmeans-numclust}
\end{figure}

For this evaluation, a logistic regression is chosen as a white-box classification model. Starting with just one cluster, the na\"ive solution of always predicting the most frequent class is achieved in both cases. Increasing the number of clusters to $k=100$ clusters, the prediction accuracy rapidly grows. By adding additional clusters, the accuracy can still be optimized further. The curve begins to flatten after 250 clusters. The final number of clusters is set to $k=400$, because for both granularity levels the prediction accuracy has somewhat converged at this point and additional clusters improve the accuracy scores only marginally. The result is a segmentation function $\varsigma(\vec{x})$, which assigns cluster labels to input vectors.

\paragraph{Company-centric Peers.}\label{Sec:Recommendation} Previous results offer a company embedding matrix $F_{K \times S'}$ of $K$ firms in a $S'=100$-dimensional space from which company distances can be derived implicitly using cosine distances. Each company $f_k$, represented by the embedding $\vec{f_k} \in F$, is indexed at position $F[k]$ in the following data structure. 

The first recommendation algorithm for peer firm identification accepts a single company, represented by its indexing position $j$ in $F$, and a fixed number $n<K$ of expected output firms. This output $f'_{1..n}$ is returned in combination with the cosine similarity scores $cos(\vec{f_\bullet}, \vec{f_j})$ to any input firm $f_j$.

\begin{center}
\begin{minipage}{0.75\textwidth}
\begin{algorithm}[H]
\caption{Identify $n$ peer firms $f'_{1..n}$ for firm $f_j$}
\begin{algorithmic}[1]
\REQUIRE company index $j$ of $f_j$ , number of expected peer firms $n<K$
\ENSURE sorted list of most similar companies $f'_{1..n}$ with $cos(\vec{f_\bullet}, \vec{f_j})$
\STATE $\vec{f_j} \leftarrow $ get company vector $F[j]$
\IF{$\vec{f_j} == \varnothing$}
\RETURN $\varnothing$
\ENDIF
\STATE $G \leftarrow \varnothing$
\FORALL{$\vec{f_k} \in F$} 
\STATE $G[k, 1]  \leftarrow $get company name $f_k  $
\STATE $G[k, 2]  \leftarrow cos(\vec{f_j}, \vec{f_k})  $
\ENDFOR
\STATE sort $G$ descending by cosine similarity $G[\ \bullet\ , 2]$
\RETURN $G[1..n, \ \bullet\ ]$
\end{algorithmic}
\end{algorithm}
\end{minipage}
\end{center}

The presented algorithm 1 initially examines the company embedding of $f_j$ and continues if such an embedding could be provided through Company2Vec. To determine $n$ similar firms, all $K$ cosine similarities between $f_j$ and $f_\bullet \in F$ are calculated and stored in $G$ with the corresponding company names. Finally, $G$ is sorted by the cosine similarity in descending order. The algorithm terminates and returns the first $n$ rows of $G$. In this formulation, the firm $f_j$ is always a peer to itself with $cos(\vec{f_j}, \vec{f_j})=1$ at the first position.

Two firms (\textit{One-vs-One}) can be compared in constant time and memory space $\mathcal{O}(1)$. A peer firm identification as proposed in algorithm 1 performs an \textit{One-vs-All} comparison, which is in $\mathcal{O}(K)$. For sorting these results, the quicksort algorithm is applied, which has an average time-complexity of $\mathcal{O}_t(K\ log\ K)$ and a space-complexity of $\mathcal{O}_s(K)$. All steps combined, algorithm 1 offers peer firms with an log-linear average time-complexity of $\mathcal{O}_t(K + K\ log\ K)  \sim  \mathcal{O}_t(K\ log\ K)$ and a linear space-complexity of $\mathcal{O}_s(1+2K)  \sim \mathcal{O}_s(K)$.

Examples of peer firms for the financial institute \textit{Commerzbank AG} are illustrated in table \ref{tab:algo1commerzbank}. The recommended peer firms all originate from the banking industry and/or offer financial services. 

\bgroup
\def\arraystretch{1.2}
\begin{table}[H] \centering 
\footnotesize
  \caption{Company-centric peer firms for \textit{Commerzbank} and n=15} 
  \label{tab:algo1commerzbank} 
\begin{tabular}{@{\extracolsep{39pt}} llc} 
\toprule
Index & Company & Similarity\\
\midrule
0   &   COMMERZBANK Aktiengesellschaft      &	1.000000\\
1   &	Degussa Bank AG	                    &	0.957322\\
2   &	Isbank AG	                        &	0.927616	\\
3   &	norisbank GmbH		                &	0.924965\\
4   &	TARGOBANK AG	                    &	0.922453	\\
5   &	CreditPlus Bank Aktiengesellschaft	&	0.918153	\\
6   &	Deutsche Kreditbank Aktiengesellschaft	&	0.916778	\\
7   &	SIGNAL IDUNA Asset Management GmbH		&	0.913773\\
8   &	Die Sparkasse Bremen AG	            &	0.902831	\\
9	&	SVS Sparkassen VersicherungsService GmbH	&	0.898554\\
10  &   EDEKABANK Aktiengesellschaft        &	0.896923\\
11	&   GLADBACHER BANK Aktiengesellschaft von 1922 & 0.896382	\\
12	&   S Broker AG \& Co. KG	            &	0.892876	\\
13	&   Sparkasse Mittelholstein Aktiengesellschaft	&	0.892515	\\
14	&   Naspa-Versicherungs-Service GmbH	&	0.891326\\
\bottomrule
\end{tabular} 
\end{table} 
\normalsize

These results reflect the accuracy of the Company2Vec model in firm-centric peer firm identification. These insights are expanded in the next paragraph by including the segmentation function.

\paragraph{Industry-centric Peers.} The first algorithm has one major drawback: It limits the output with a fixed cut-off value to $n$. As some companies encounter a varying number of potential peer firms, this number is in the following identified fully by the Company2Vec model using the pretrained segmentation function $\varsigma(\vec{x})$ of subsection \ref{Sec:ClusteringFunction}.

\begin{center}
\begin{minipage}{0.75\textwidth}
\begin{algorithm}[H]
\caption{Identify all $N$ peer firms $f'_{1..N}$ for firm $f_{j}$}
\begin{algorithmic}[1]
\REQUIRE company index $j$ of $f_j$ , pretrained segmentation function $\varsigma(\vec{x})$ 
\ENSURE list of all similar companies $f'_{1..N}$
\STATE $\vec{f_j} \leftarrow $ get company vector $F[j]$
\IF{$\vec{f_j} == \vec{0}$}
\RETURN $\varnothing$
\ENDIF
\STATE $s_j \leftarrow \varsigma(\vec{f_j})$ 
\STATE $G \leftarrow [\varnothing$]
\FORALL{$\vec{f_k} \in F$} 
\STATE $s_k \leftarrow \varsigma(\vec{f_k})$ 
\IF{$s_k == s_j$}
\STATE $G \leftarrow G \cup f_k$
\ENDIF
\ENDFOR
\RETURN $G$
\end{algorithmic}
\end{algorithm}
\end{minipage}
\end{center}

For a company $f_j$ with a non-empty company embedding $\vec{f_j}$, the pretrained clustering function assigns a cluster label $s_j$ to this firm. Afterwards, the algorithm iterates over the list of all $K$ companies and stores those companies that belong to the same cluster in the output list $G$. 

The use of a pretrained segmentation function offers a linear cluster assignment of a single company in $\mathcal{O}_t(1)$. Consequently, iterating over all companies and identifying peer firms is performed with $\mathcal{O}_t(K)$ in linear time, since no additional sorting is applied. In a worst-case scenario, where all companies are clustered into one single cluster, the space complexity is still linear with  $\mathcal{O}_s(2+2K) \sim \mathcal{O}_s(K)$. Hence, the algorithm 2 is more efficient than the first one.

\bgroup
\def\arraystretch{1.2}
\begin{table}[H] \centering 
\footnotesize
  \caption{Industry-centric peer firms for \textit{Microsoft}, \textit{Vattenfall} and \textit{K\&D Hausbau}.} 
  \label{tab:industry-centric} 
\begin{tabular}{@{\extracolsep{5pt}} lll} 
\toprule
Microsoft  & Vattenfall & K\&D Hausbau\\
\midrule
Artwizz 	                    &	BTB-Blockheizkraftwerks...	                &	Manos Grundstücksv…	\\
AVM Computersysteme 	        &	Thermische Abfallbehandl...	                &	Peterson Warenhandels…	\\
ony Europe B.V.                 &	SES Energiesysteme 	                            &	Krynos Einfamilienhaus	\\
KOSATEC Computer 	            &	Müllverwertung Borsig...	                &	petruswerk Grundbesitz	\\
Supplies Hard- u. Software...	&	Danpower 	                                    &	ZET-Bauträgergesellschaft	\\
everIT 	                        &	Danpower Energie Service 	                    &	TAG Wohnungsgesellsch…	\\
Intenso International 	        &	EEW Energy from Waste A...   	                &	TAG Wohnen 	\\
Meinberg Funkuhren 	            &	IHKW Industrieheizkraft...	                &	LUKAS Massivhaus \\
CSL-Computer 	                &	Entsorgungszentrum Salzg...	                &	Odinstraße Projektentw…	\\
Handy Deutschland 	            &	EBS Kraftwerk 	                                &Bonava Wohnimmobilien	\\
MEV Elektronik Service 	        &	EEW Energy from Waste P...	                &	Bonava Deutschland 	\\
ACER Computer 	                &	INGAVER Innovative Gas...	                &	Bonava Wohnbau	\\
Delta Computer Products 	    &	Harzwasser - K. Wasserv...	                &	Fuchs Massivhaus	\\
CASIO Europe 	                &	Harzwasserwerke 	                &	Nibelungen-Wohnbau	\\
VITREX Multimedia 	            &	EnviTec Anlagenbau                   &	ASSMANN Beraten...
	\\

\multicolumn{1}{c}{$\vdots$} & \multicolumn{1}{c}{$\vdots$}&\multicolumn{1}{c}{$\vdots$} \\

\midrule
\multicolumn{1}{c}{65 peer firms} & \multicolumn{1}{c}{145 peer firms} &\multicolumn{1}{c}{280 peer firms} \\

\bottomrule
\end{tabular} 
\end{table} 
\normalsize

The following example demonstrates recommended industry-centric peer firms for three random firms from different industries. A sample of 15 firms is drawn out of each cluster with homogeneous business activities (table \ref{tab:industry-centric}).

The recommended firms of algorithm 2 demonstrate the accuracy of Company2Vec in industry-centric peer firm identification.

For instance, typical peer firms of \textit{Microsoft} are technology companies with a focus on hardware production, software development or other IT services. In total, 65 peer firms are proposed in this cluster. The energy producer \textit{Vattenfall} belongs to a cluster with 145 peer firms, mostly companies from energy or water supply and waste recycling. The third company, which belongs to the largest cluster in this example, \textit{Kirsch \& Drechsler Hausbau}, is a home construction company and is accompanied by 280 firms from the construction and living sector.

\paragraph{Portfolio-centric Peers.} A vectorized representation of companies allows the identification of peer firms based on a firm portfolio with $m$ firms. For this purpose, algorithm 3 proposes the same vector averaging technique as previously used for topic modeling, which has proved successful on company data. Hence, each company vector of the portfolio is summed up and then divided by the portfolio size. The resulting portfolio vector $\vec{p}$ is treated as a single company that combines all business activities of the portfolio. Finally, the procedure follows the same steps as algorithm 1 for identifying peer firms based on this portfolio vector. 

In large or diversified portfolios, the averaging methods put more weight on larger shares of firms that form homogeneous groups of business activities within the portfolio. Hence, the information of a single outlying firm is included in the portfolio vector with a share of only $\frac{1}{m}$. Hence, for larger portfolios with mostly homogeneous business activities, the problem of outlying firms can be omitted. Multiple intra-portfolio business activities are represented by forming a geometric center of all different activities.

The complexity analysis of this algorithm is straightforward: The portfolio vector is calculated in linear time ($\mathcal{O}_t(m)$) with constant memory space ($\mathcal{O}_s(1)$). The subsequent peer firm identification follows the same complexity as algorithm 1. In summary, algorithm 3 has an average run-time complexity of $\mathcal{O}_t(m + K\ log\ K) \sim \mathcal{O}_t(K\ log\ K)$ for $m\leq K$ and a memory space complexity of $\mathcal{O}_s(2+2K) \sim \mathcal{O}_s(K)$.

\begin{center}
\begin{minipage}{0.75\textwidth}
\begin{algorithm}[H]
\caption{Identify $n$ peer firms $f'_{1..n}$ for portfolio with $m$ firms $f_{1..m}$.}
\begin{algorithmic}[1]
\REQUIRE company indices $j_{1..m}$ of $f_{j_{1..m}}$ , number of expected peer firms $n$
\ENSURE list of all similar companies $f'_{1..n}$
\STATE $\vec{p} \leftarrow \vec{0}$
\FORALL{$f_k \in f_{j_{1..m}}$} 
\STATE $\vec{f_k} \leftarrow $ get company vector $F[k]$
\STATE $\vec{p} \leftarrow \vec{p} + \vec{f_k}$
\ENDFOR
\STATE $\vec{p} \leftarrow m^{-1} \cdot \vec{p}$
\IF{$\vec{p} == \varnothing$} 
\RETURN $\varnothing$
\ENDIF
\STATE $G \leftarrow \varnothing$
\FORALL{$\vec{f_{k'}} \in F$} 
\STATE $G[k', 1]  \leftarrow$ get company name $f_{k'} $
\STATE $G[k', 2]  \leftarrow cos(\vec{p}, \vec{f_{k'}})  $
\ENDFOR
\STATE sort $G$ descending by cosine similarity $G[\ \bullet\ , 2]$
\RETURN $G[1..n, \ \bullet\ ]$
\end{algorithmic}
\end{algorithm}
\end{minipage}
\end{center}

To present an understandable illustration, the number of firms in the following portfolio is set to $m=2$. The financial institute \textit{Commerzbank} and the home construction company \textit{Kirsch \& Drechsler Hausbau} are used as a diversified two-firm portfolio for which similar firms are identified. 
The joint cosine similarity of these two firms is 0.2728 and they can thus be interpreted as somewhat different. The business activities of a bank and a construction company are not directly related to each other.

Table \ref{tab:portfolio-centric} demonstrates that Company2Vec is able to form accurate \textit{semantic averages} of both industries. The semantic combination of a bank and a home construction company is identified as \textit{building society} or \textit{mortgage bank}, which is reflected in the output of the $n=15$ peer firms.

\bgroup
\def\arraystretch{1.2}
\begin{table}[H] \centering 
\footnotesize
  \caption{Portfolio-centric peer firms for \textit{Commerzbank} and \textit{K\&D Hausbau} and n=15.} 
  \label{tab:portfolio-centric} 
\begin{tabular}{@{\extracolsep{36pt}} llc} 
\toprule
Index & Company & Similarity\\
\hline
0	&	Wüstenrot Immobilien GmbH	&	0.889722	\\
1	&	Wüstenrot Bausparkasse Aktiengesellschaft	&	0.889078	\\
2	&	Bausparkasse Schwäbisch Hall Aktiengesellschaft	&	0.880761	\\
3	&	Sparkasse zu Lübeck Aktiengesellschaft	&	0.879065	\\
4	&	LBS Immobilien GmbH	&	0.872263	\\
5	&	IFS GmbH Immofinanz Sachwertanlagen	&	0.870594	\\
6	&	LBS Bausparkasse Schleswig-Holstein-Hamburg AG	&	0.862315	\\
7	&	Stille Immobilien GmbH \& Co. KG	&	0.862127	\\
8	&	SIGNAL IDUNA Bauspar Aktiengesellschaft	&	0.860815	\\
9	&	Garant Immobilien Heilbronn GmbH \& Co. KG	&	0.858831	\\
10	&	AMADEUS Projektbau GmbH	&	0.850665	\\
11	&	DKB Wohnimmobilien Beteiligungs GmbH \& Co. KG	&	0.850474	\\
12	&	I.W.E.S. GmbH \& Co. KG	&	0.849096	\\
13	&	INTERHYP AG	&	0.848369	\\
14	&	neue leben Pensionskasse Aktiengesellschaft	&	0.845932	\\
\bottomrule
\end{tabular} 
\end{table} 
\normalsize

A TSNE visualisation of different German companies underlines the semantic structures of the company embeddings which are important for applying the averaging method (see appendix, Fig. \ref{Fig:TSNE-examples}).

\section{Discussion and Conclusion}\label{Sec:Discussion}
In summary, Company2Vec offers an innovative approach for creating efficient company embeddings on textual and visual data derived from company webpages. The study shows that topic modeling with Word2Vec on website data is able to reflect companies' business activities.
The embeddings can be used in a supervised evaluation setting for industry prediction. In terms of accuracy, the classification results come close to human-like classification results (subsection \ref{Par:compEmbEval}). The study shows that true industry structures are reflected through semantic modeling (subsection \ref{Sec:Semantics}). Considering unsupervised peer firm identification, three algorithms identify (1) \textit{firm-centric}, (2) \textit{industry-centric} and (3) \textit{portfolio-centric} peer firms (subsection \ref{Sec:Recommendation}). This study was the first to successfully apply \textit{firm-centric} and \textit{industry-centric} peer firm identification on textual and visual company webpage data. The \textit{portfolio-centric} approach can determine semantic averages of multiple firms and has not yet been proposed in literature (table \ref{tab:portfolio-centric}).  

However, there are some limitations of this study's design and the results. With a successful webscraping of about $92\%$ of the original URL list, the majority of companies could be included in the \textbf{Company2Vec database}. Due to anti-scraping tools, the remaining $8\%$ are blocked and cannot be analysed by the model.
Furthermore, as highlighted in the literature, \textbf{evaluation sets} with the true industry labels also contain misclassifications. Hence, the data cannot be interpreted as correctly for all firms.\cite{Pierre2001OnTA} This restriction also applies to the evaluation set of this study.  

The final process of data transformation uses a \textbf{pretrained Word2Vec model} with a high correlation to empirical semantic word-pair relations. 
It is not possible to add additional training epochs to the model by Müller \cite{mueller2015}. Thus, the supplementary company vocabulary is not included.  

Comparing information gain using \textbf{textual-vs-visual data} clearly states that the textual information on company webpages is the most important factor. Only in fine-granular industry segmentation, the visual data is able to produce marginally more accurate results. One reason lies within the fact that a significant portion of companies (about $25\%$) does not have images on their company webpages. An additional reason is the image classifier, which has to be improved in further research.

On the one hand, Company2Vec is able to identify similar firms based on their business activities. On the other hand, \textbf{additional features} such as company size or headquarters are not included. In the present setting, this could only be achieved by applying an additional filter to the final company recommendations. Hence, auxiliary data is needed for such filter specifications. 

The company segmentation is performed with k-means clustering. This algorithm is able to form well-proportioned segments of firms. One drawback of this approach is that k-means is \textbf{non-deterministic}. Therefore, the exact cluster assignment cannot be reproduced on different data or with a different number of clusters.   

The three presented algorithms for peer firm identification are analysed in terms of \textbf{run-time and memory space complexity}. Algorithms (1) and (3) both implement a sorting algorithm that has a log-linear run-time complexity of $\mathcal{O}_t(K\ log\ K)$. Especially for a smaller number of output firms $n$ and a large company database $K$, a selection algorithm - instead of complete sorting - could improve the performance of peer firm identification. Hence, the run-time complexity could be reduced to $\mathcal{O}_t(n\cdot K) \sim \mathcal{O}_t(K)$ instead of $\mathcal{O}_t(K\ log\ K)$ by iterating $n$ times over $K$ with $n\ll K$ and thus returning the firm with the maximum cosine similarity each time. \\

This study made use of modern machine-learning techniques to develop Company2Vec: a method for efficient company representation based on unstructured textual and visual data from German company webpages. Applying topic modeling approaches, the final procedure uses a pretrained Word2Vec model and creates accurate company embeddings that represent the business activities of firms. Both textual and visual data prove to be relevant for fine-granular distinctions between companies. 

The \textit{recommender algorithms} (1-3) identify peer firms accurately with low or moderate run-time and memory space complexity. Thus, the Company2Vec model provides efficent peer firm identification using (1) \textit{firm-centric}, (2) \textit{industry-centric} and (3) \textit{portfolio-centric} approaches.

The insights and conclusions of this work can be transferred to different company data using the URL of new corporate webpages. In the future, it would be of interest to include structured data, such as financial balance sheets, company size or company location to the Company2Vec model. The development of enhanced company embeddings that combine business activities and additional information might also require more sophisticated recommender algorithms. In particular, deep neural networks as proposed by Covington et al. \cite{10.1145/2959100.2959190} for YouTube recommendations might outperform algorithms (1-3) on certain tasks. This is just one example of the challenges ahead for further company similarity research.

 \section*{Appendices}\label{Sec:appendix}

\bgroup
\def\arraystretch{1.2}
\begin{table}[H] \centering 
\footnotesize
  \caption{Evaluation of pretrained and self-trained Word2Vec models.} 
  \label{tab:word2vecs} 
\begin{tabular}{@{\extracolsep{3pt}} lccccc} 
\toprule
& \multirow{2}{*}{Dimensions} & \multicolumn{2}{c}{WordSim-353} & \multicolumn{2}{c}{SimLex-999}\\
 \cline{3-4}  \cline{5-6}
 \vspace{2px}
& &  Correlation & Coverage & Correlation & Coverage \\
\midrule
Müller et al. (2015)\cite{mueller2015} &300& 0.5861 & \textbf{0.9717} & \textbf{0.3678} & \textbf{0.9750} \\
Fares et al. (2017)\cite{fares-etal-2017-word} &100& 0.5821 & 0.9660 & 0.2930 & 0.9630 \\
Yamada et al. (2018)\cite{Wikipedia2Vec2018} &300& \textbf{0.5955} & 0.8640 & 0.3496 & 0.8168 \\
Self-trained A (2021)\scriptsize{*} &100& 0.0761 & 0.8159 & 0.0799 & 0.7778 \\
Self-trained B (2021)\scriptsize{*} &300& 0.0609 & 0.8159 & 0.0412 & 0.7778 \\
\bottomrule
\end{tabular} 
\end{table} 
\vspace{-15mm}
\scriptsize{\hspace{20mm}* trained only on company webpage data corpus}\\
\normalsize

\begin{figure}[H]
\centering
\footnotesize
\begin{tikzpicture}
  \node (img1)  {  \includegraphics[trim=95 90 300 100, clip,height=5cm]{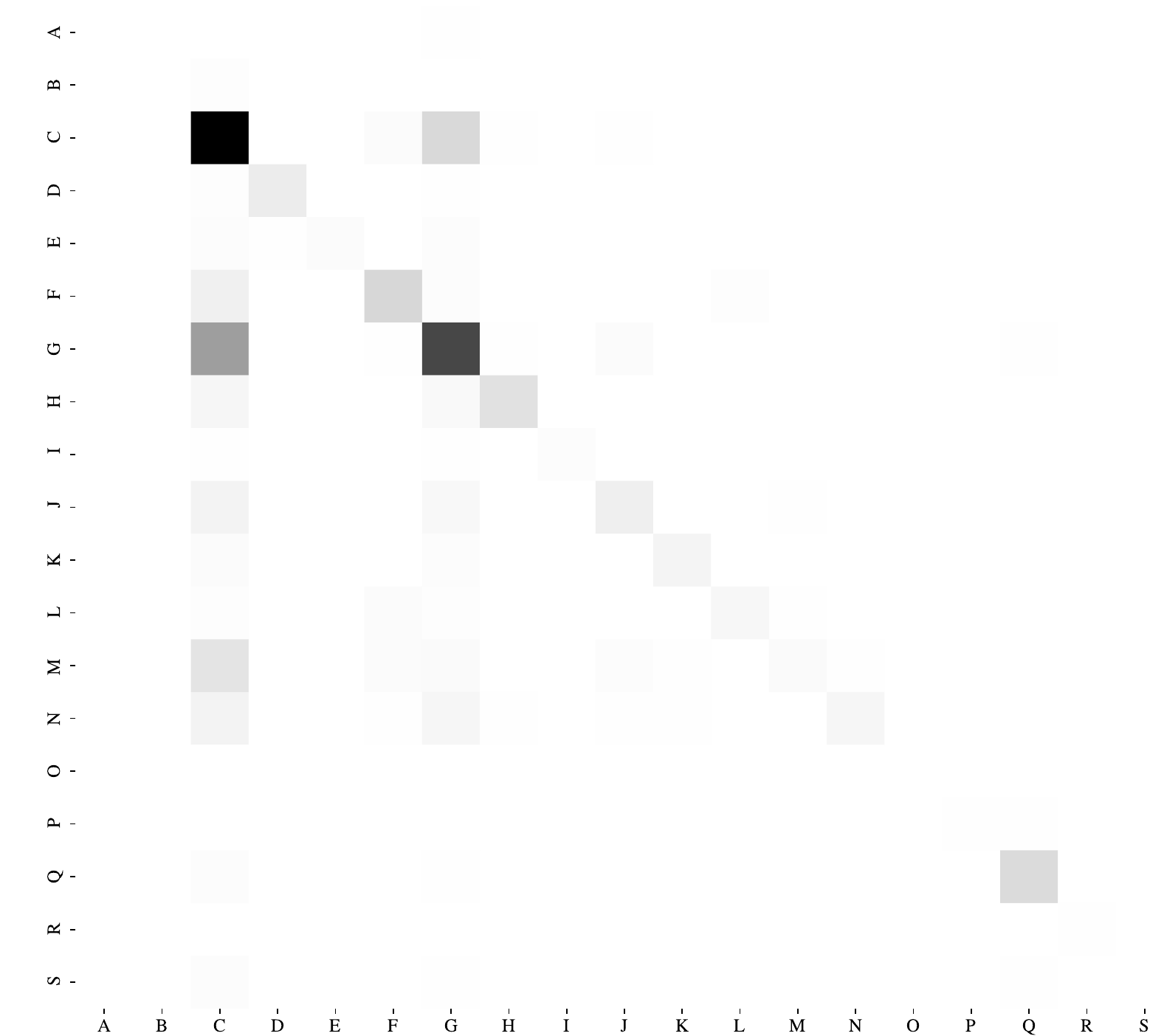}};
  \node[below=of img1, node distance=0cm, yshift=0.6cm,font=\color{black}] {predicted class};
  \node[above=of img1, node distance=0cm,font=\color{black}] {\textbf{Imbalanced Level-1 Industries}};  
  \node[left=of img1, node distance=0cm, rotate=90, anchor=center,font=\color{black}] {true class};  
  \node[right=of img1,xshift=0cm, yshift=0cm] (img2)  {\includegraphics[trim=95 90 100 100, clip,height=5cm]{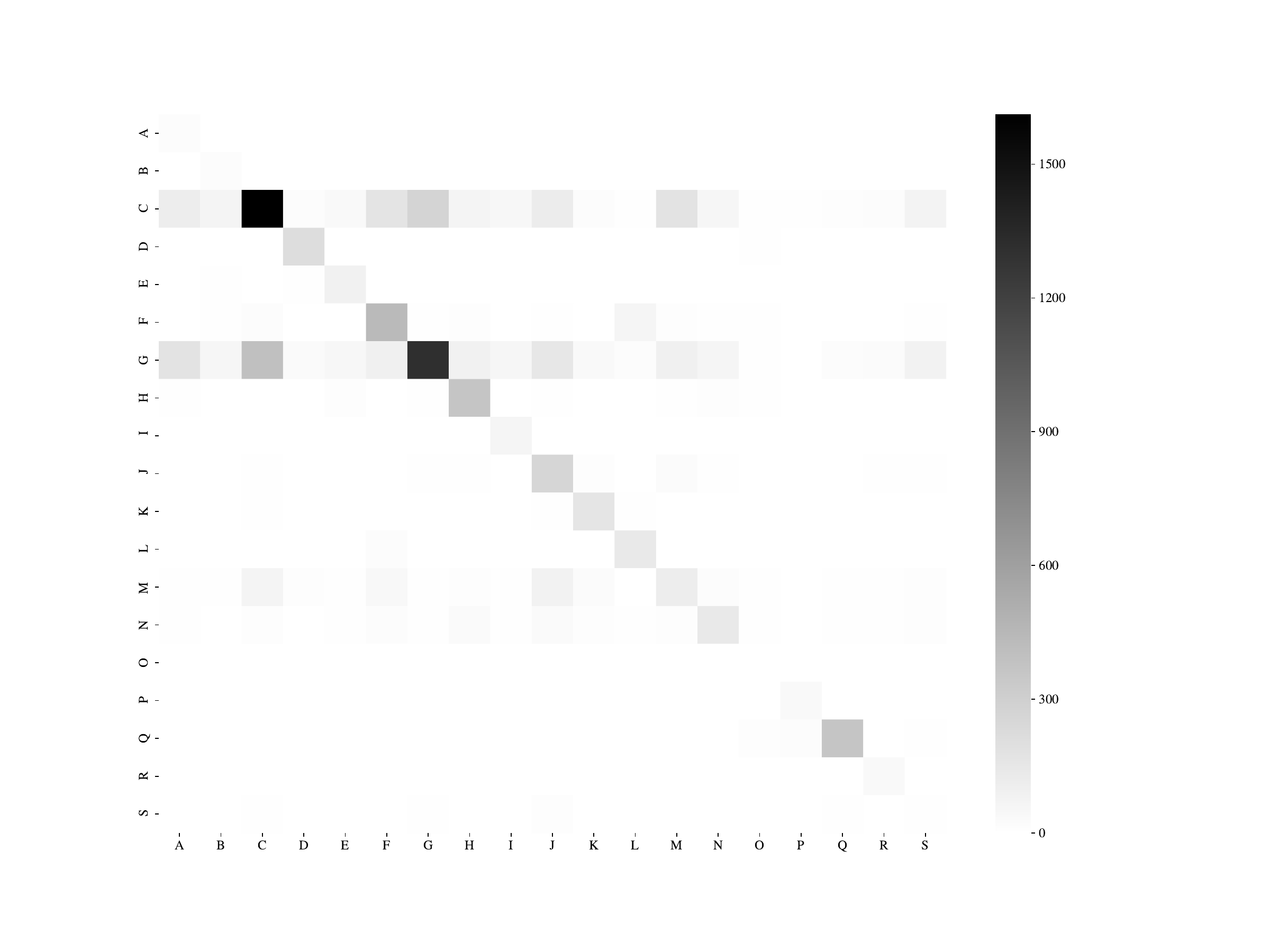}};
  \node[below=of img2, node distance=0cm, xshift=-0.5cm, yshift=0.6cm, font=\color{black}] {predicted class};  
  \node[above=of img2, node distance=0cm, xshift=-0.5cm,font=\color{black}] {\textbf{Balanced Level-1 Industries}};  
\end{tikzpicture}
\caption{Confusion matrices of level-1 industry prediction task for Word2Vec embeddings.}
\label{Fig:conf_level1-w2v}
\end{figure}
\normalsize

\begin{figure}[H]
\centering
\footnotesize
\begin{tikzpicture}
  \node (img1)  {  \includegraphics[trim=95 90 300 100, clip,height=5cm]{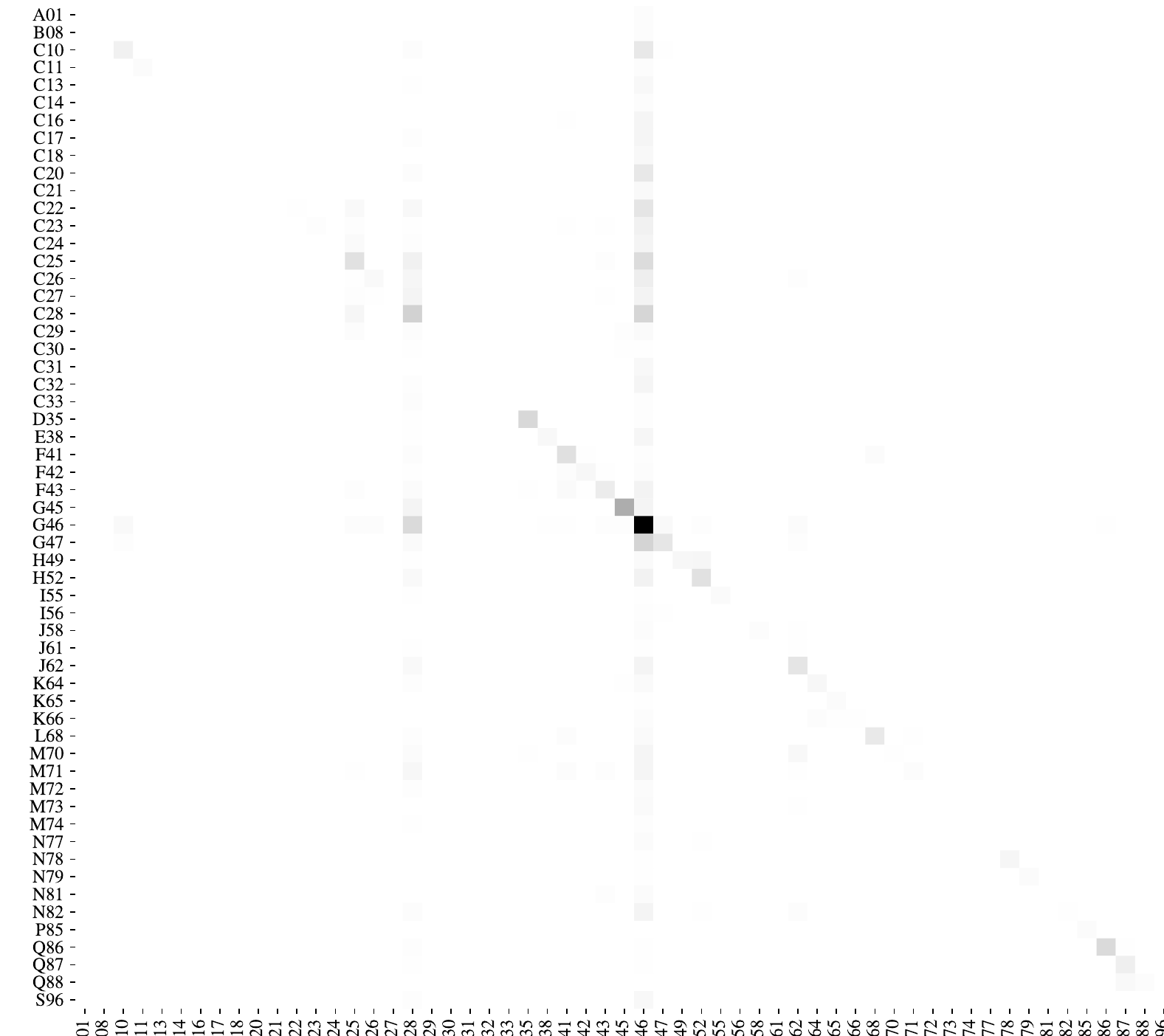}};
  \node[below=of img1, node distance=0cm,font=\color{black}] {predicted class};
  \node[above=of img1, node distance=0cm,font=\color{black}] {\textbf{Imbalanced Level-2 Industries}};
  \node[left=of img1, node distance=0cm, rotate=90, anchor=center,font=\color{black}] {true class};
  \node[right=of img1, yshift=0cm] (img2)  {\includegraphics[trim=95 90 100 100, clip,height=5cm]{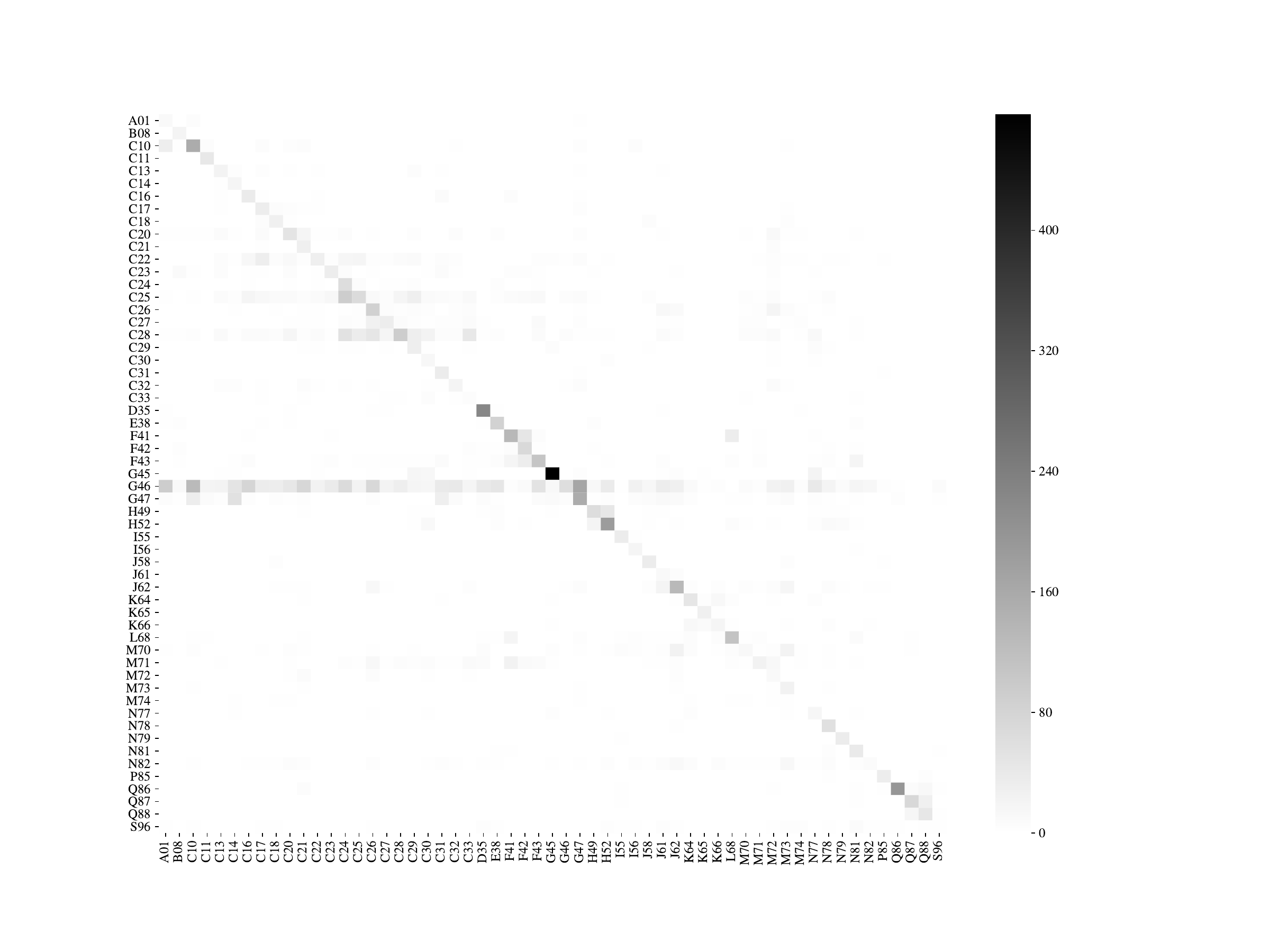}};
  \node[below=of img2, node distance=0cm, xshift=-0.5cm,font=\color{black}] {predicted class};
 \node[above=of img2, node distance=0cm, xshift=-0.5cm,font=\color{black}] {\textbf{Balanced Level-2 Industries}};
\end{tikzpicture}
\caption{Confusion matrices of level-2 industry prediction task for Word2Vec embeddings.}
\label{Fig:conf_level2-w2v}
\end{figure}
\normalsize

\begin{figure}[H]
\centering
\includegraphics[trim=95 90 110 100, clip,width=1\textwidth]{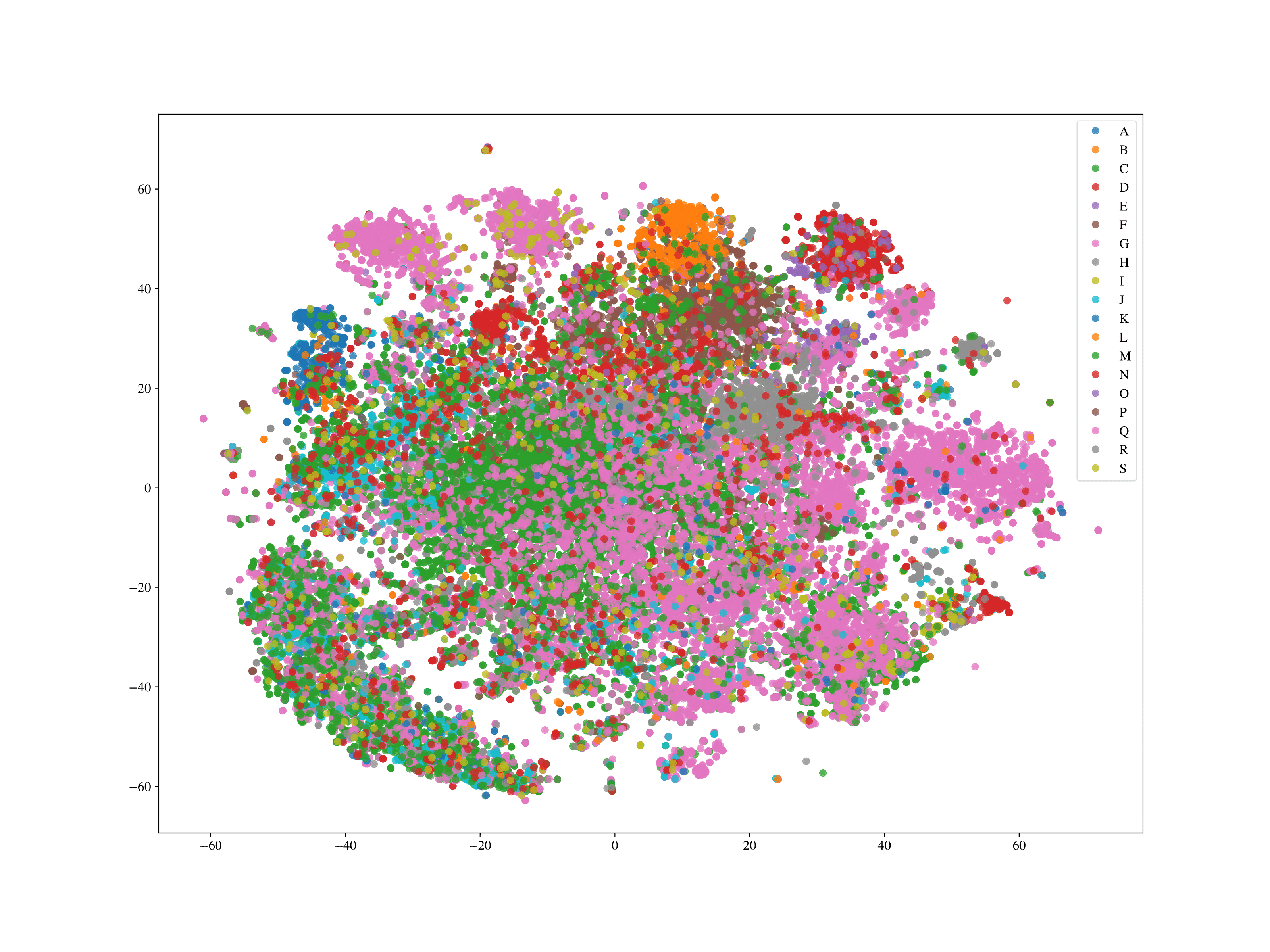}
\caption{TSNE plot of industry structure with original level-1 industry classification.}
\label{Fig:TSNE-19}
\end{figure}

\begin{figure}[H]
\centering
\includegraphics[trim=95 30 110 50, clip,width=1\textwidth]{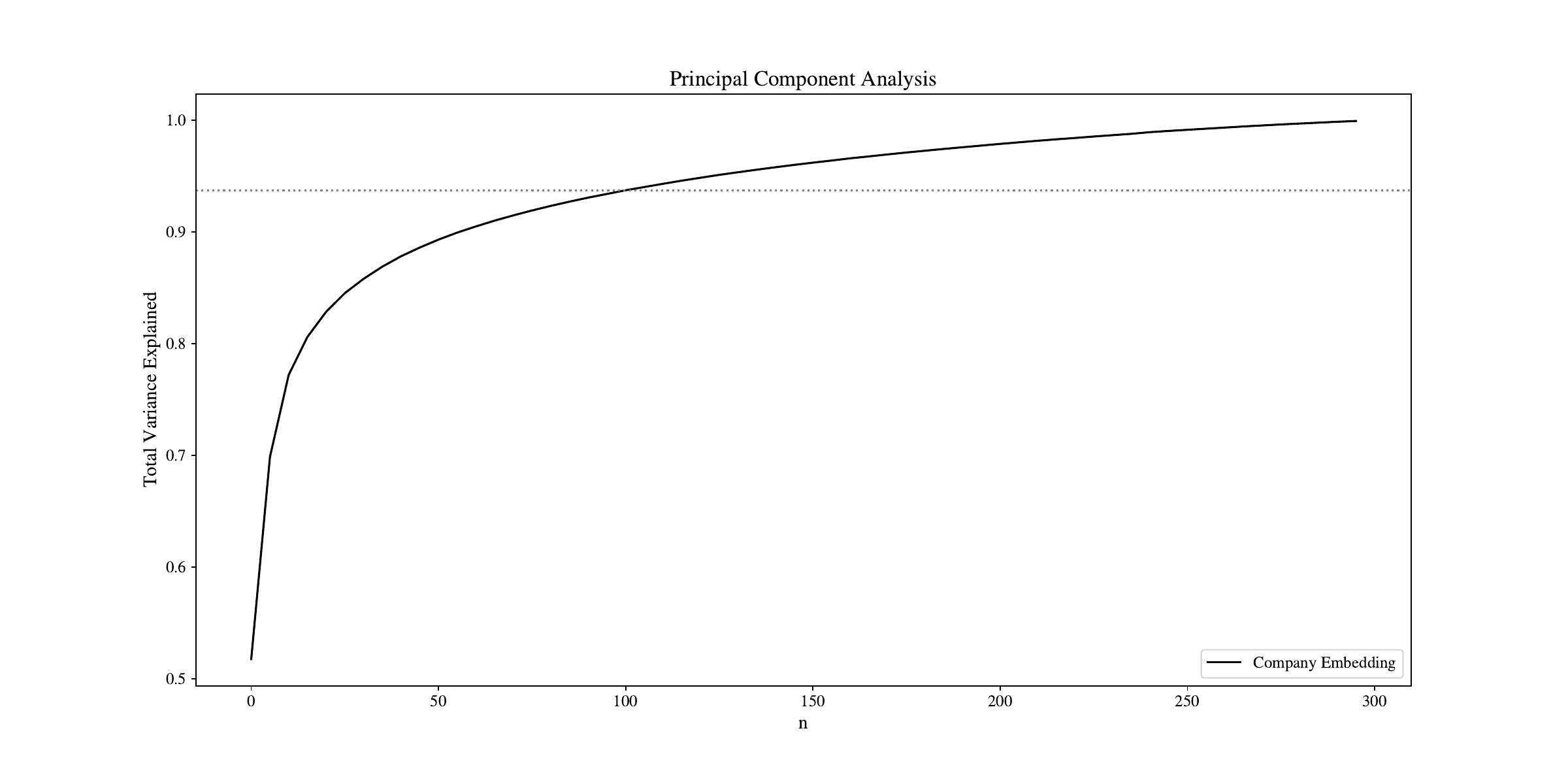}
\caption{Dimensionality reduction of 300-dimensional Word2Vec company embeddings.}
\label{Fig:pca}
\end{figure}

\begin{figure}[H]
\centering
\includegraphics[trim=110 90 110 100, clip,width=1\textwidth]{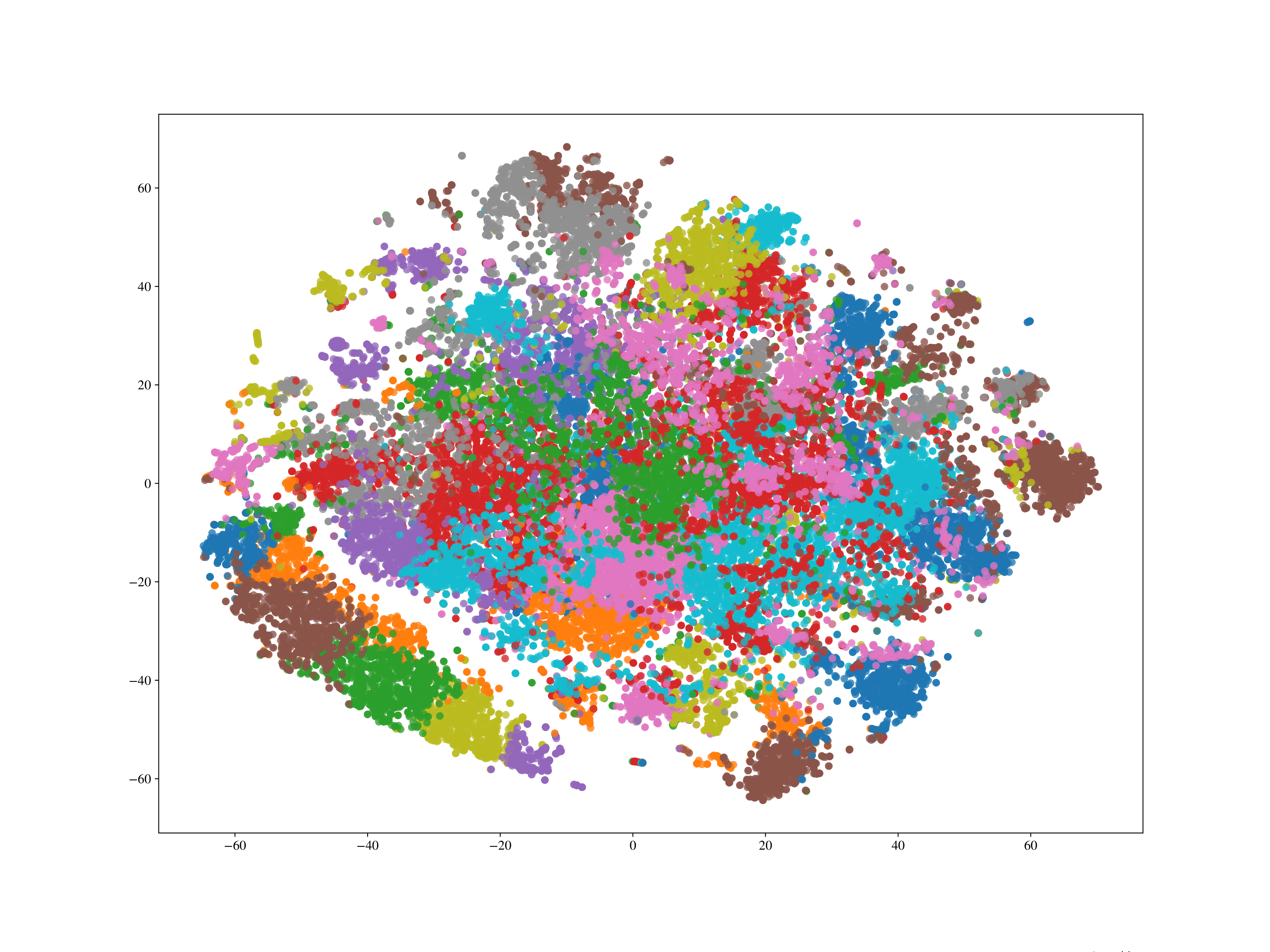} 
\caption{TSNE plot (10 colours only) of k=50 clusters using k-means.}
\label{Fig:TSNE-kmeans-50}
\end{figure}

\begin{figure}[H]
\centering
\includegraphics[trim=90 30 110 50, clip,width=1\textwidth]{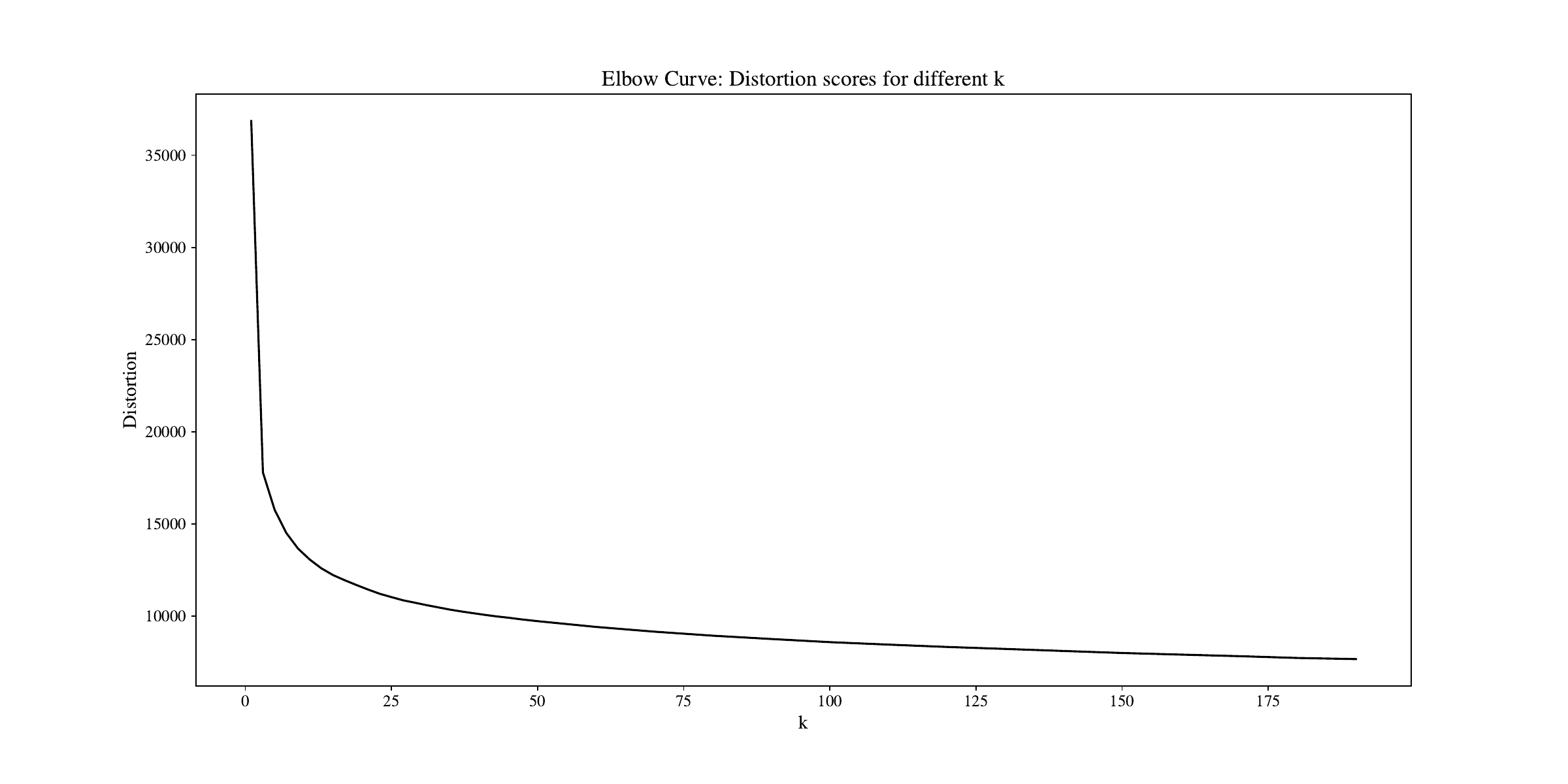}
\caption{Distortion for different k in k-means clustering.}
\label{Fig:kmeans-elbow}
\end{figure}

\bibliographystyle{IEEEtran}   
\bibliography{00_literature}

\end{document}